\documentclass[times,twocolumn,final,authoryear]{elsarticle}

\usepackage{framed,multirow}

\usepackage{amssymb}
\usepackage{latexsym}

\usepackage{url}
\usepackage{xcolor}
\definecolor{newcolor}{rgb}{.8,.349,.1}

\usepackage[colorlinks]{hyperref}
\usepackage{graphicx}
\usepackage{amssymb}
\usepackage{amsmath}
\usepackage{pifont}
\usepackage{array}
\usepackage{multirow}
\usepackage{hyphenat}
\usepackage{booktabs}
\usepackage{colortbl}
\usepackage{arydshln}
\usepackage{soul}
\usepackage{tabularx}
\usepackage{lscape}
\usepackage[figuresleft]{rotating}
\usepackage[switch]{lineno}

\usepackage{caption}
\usepackage[labelformat=simple]{subcaption}

\usepackage{xspace}
\makeatletter
\DeclareRobustCommand\onedot{\futurelet\@let@token\@onedot}
\def\@onedot{\ifx\@let@token.\else.\null\fi\xspace}

\def\eg{\emph{e.g}\onedot} 
\def\ie{\emph{i.e}\onedot} 
 
\def\etc{\emph{etc}\onedot} 
 
\def\etal{\emph{et al}\onedot}
\makeatother

\usepackage[capitalize]{cleveref}
\crefname{section}{Sec.}{Secs.}
\Crefname{section}{Section}{Sections}
\Crefname{table}{Table}{Tables}
\crefname{table}{Tab.}{Tabs.}
\Crefname{figure}{Figure}{Figures}
\crefname{figure}{Fig.}{Figs.}
\Crefname{equation}{Equation}{Equations}
\crefname{equation}{Eq.}{Eqs.}

\usepackage{caption}
\usepackage{subcaption}
\usepackage{enumitem}
\setlist[itemize]{noitemsep,nolistsep}

\newcolumntype{C}{>{\centering\arraybackslash}X}
\newcolumntype{L}{>{\raggedright\arraybackslash}X}
\newcolumntype{R}{>{\raggedleft\arraybackslash}X}

\newcommand{\cmark}{\ding{51}}
\newcommand{\xmark}{\ding{55}}

\newcommand{\red}[1]{\textcolor{red}{#1}}

\newcommand{\para}[1]{\vspace{3pt}\noindent\textbf{#1}\xspace}

\newcommand{\Rn}[1]{\uppercase\expandafter{\romannumeral#1}}

\newcommand{\mr}[1]{\mathrm{#1}}
\newcommand{\mc}[1]{\mathcal{#1}}
\newcommand{\norm}[1]{\left\lVert #1 \right\rVert_2}

\DeclareMathOperator*{\argmax}{arg\,max}

\graphicspath{{./images/}{./rev_images/}}

\begin{document}

\ifpreprint
  \setcounter{page}{1}
\else
  \setcounter{page}{1}
\fi

\begin{frontmatter}

\title{Semi-supervised Cycle-GAN for face photo-sketch translation in the wild}


\author[1]{Chaofeng Chen} 
\author[2]{Wei Liu}
\author[3]{Xiao Tan}
\author[1]{Kwan-Yee~K. Wong}

\address[1]{The Department of Computer Science, The University of Hong Kong,
Pokfulam Road, Hong Kong SAR, China}
\address[2]{SenseTime Research, Shenzhen 518000, China}
\address[3]{Baidu Inc, Department of Computer Vision Technology, Beijing 100085, China.}


\begin{abstract}

The performance of face photo-sketch translation has improved a lot thanks to deep neural networks. GAN based methods trained on paired images can produce high-quality results under laboratory settings. Such paired datasets are, however, often very small and lack diversity. Meanwhile, Cycle-GANs trained with unpaired photo-sketch datasets suffer from the \emph{steganography} phenomenon, which makes them not effective to face photos in the wild. In this paper, we introduce a semi-supervised approach with a noise-injection strategy, named Semi-Cycle-GAN (SCG), to tackle these problems. For the first problem, we propose a {\em pseudo sketch feature} representation for each input photo composed from a small reference set of photo-sketch pairs, and use the resulting {\em pseudo pairs} to supervise a photo-to-sketch generator $G_{p2s}$. The outputs of $G_{p2s}$ can in turn help to train a sketch-to-photo generator $G_{s2p}$ in a self-supervised manner. This allows us to train $G_{p2s}$ and $G_{s2p}$ using a small reference set of photo-sketch pairs together with a large face photo dataset (without ground-truth sketches). For the second problem, we show that the simple noise-injection strategy works well to alleviate the \emph{steganography} effect in SCG and helps to produce more reasonable sketch-to-photo results with less overfitting than fully supervised approaches. Experiments show that SCG achieves competitive performance on public benchmarks and superior results on photos in the wild. 

\end{abstract}



\end{frontmatter}


\section{Introduction}\label{sec:intro}
Face photo-sketch translation can be considered as a specific type of image translation between an input face photo and sketch. It has a wide range of applications. For example, police officers often have to identify criminals from sketch images, sketch images are also widely used in social media. 

\begin{figure}[t]
\newcommand{\imgwidth}{0.19\linewidth}
\centering
\begin{subfigure}[b]{\linewidth}
\includegraphics[width=\linewidth]{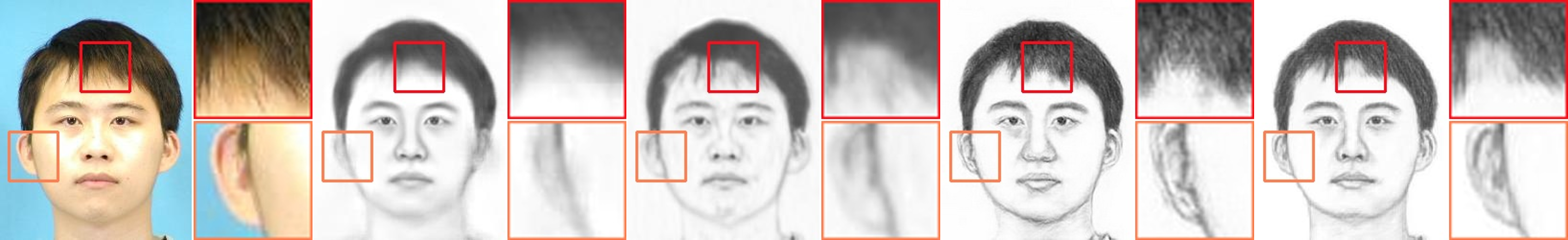}
\\
\makebox[\imgwidth]{\scriptsize (i) Photo}
\makebox[\imgwidth]{\scriptsize (ii) RSLCR}
\makebox[\imgwidth]{\scriptsize (iii) FCN}
\makebox[\imgwidth]{\scriptsize (iv) Pix2Pix}
\makebox[\imgwidth]{\scriptsize (v) Ours}
\caption{Example results of different methods on the public benchmarks.} \label{fig:intro}
\end{subfigure}
\begin{subfigure}[b]{\linewidth}
    \includegraphics[width=\linewidth]{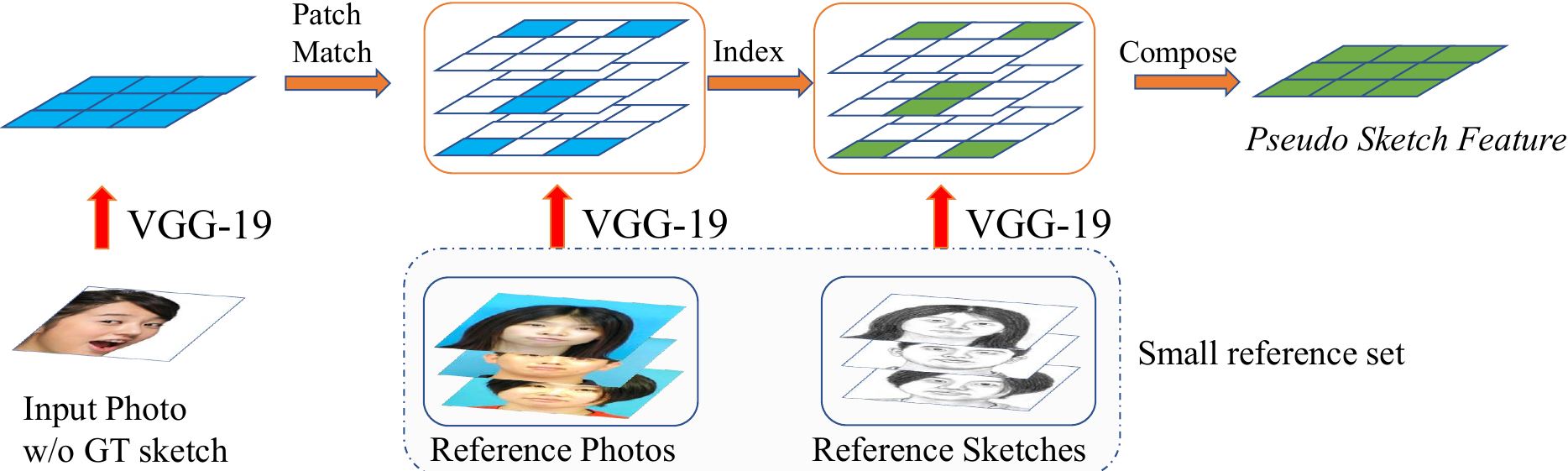}
    \caption{Pipeline for constructing the \emph{pseudo sketch feature} (PSF)} \label{fig:intro_psf}
\end{subfigure}
\caption{Example results comparison and the proposed pseudo sketch feature.}
\vspace{-1em}
\end{figure}

There are lots of works on face photo-sketch translation. Traditional methods are based on patch matching. They usually divide an input photo into small patches and find corresponding sketch patches in a reference dataset composed of well-aligned photo-sketch pairs. In this way, they \citep{song2014real,zhou2012markov,ijcai2017-500,wang2009face} achieved pleasant results without explicitly modeling the  mapping between photos and sketches, which is highly non-linear and difficult. However, sketches generated by these methods are often over-smoothed and lack subtle contents, such as ears in \cref{fig:intro}(ii). Moreover, these methods are usually very slow due to the time-consuming patch matching and optimization process. Recent methods based on Convolutional Neural Networks (CNNs) try to directly learn the translation between photos and sketches. However, results produced by simple CNNs are usually blurry (see \cref{fig:intro}(iii)), and Generative Adversary Networks (GAN) \citep{goodfellow2014generative} often generate unpleasant artifacts (see \cref{fig:intro}(iv)). Finally, due to the lack of large training datasets, these learning-based approaches cannot generalize well to photos in the wild. 

Latest works \citep{Gao2017cagan,Wang2017psman,FANG-IACycleGAN} utilize Cycle-GAN \citep{CycleGAN2017} to learn the translation between photos and sketches. Cycle-GAN is designed for unpaired translation between different domains. Styles are translated with a discriminator loss and content consistency is guaranteed with a cycle-consistency loss. However, the cycle-consistency loss used to constrain content is weak, and therefore these methods still require paired data to calculate an MSE (mean squared error) loss between the prediction and ground truth. In experiments, we observed that models directly using unpaired Cycle-GAN fail to preserve facial content (see \cref{fig:footprint}). This is because Cycle-GAN learns to “hide” information of the input photos in the generated sketches as invisible high-frequency noise, also called \emph{steganography} \citep{chu2017steganography,selfdefense}. It makes it difficult to learn face photo-sketch translation with Cycle-GAN in an unpaired setting. Please refer to \cref{sec:review-cyclegan} for a detailed discussion.

In this paper, we propose a semi-supervised learning framework based on Cycle-GAN, named Semi-Cycle-GAN (SCG), for face photo-sketch translation. To ensure content consistency, we introduce a novel {\em pseudo sketch feature} (PSF) to supervise the training of the photo-to-sketch generator $G_{p2s}$. \Cref{fig:intro_psf} shows the pipeline to construct PSF for an input photo without ground truth sketch. Suppose we have a small reference set of photo-sketch pairs and a large face photo dataset without ground-truth sketches. Similar to the exemplar-based approach, we first subdivide an input photo and its VGG-19 \citep{simonyan2014very} feature maps into overlapping patches. We then match (in the feature space) these photo patches with the photo patches in the reference set and compose a PSF from the VGG-19 features of the corresponding sketch patches in the reference set. We next supervise the training of $G_{p2s}$ using the MSE between the feature maps of the generated sketch and the PSF of the input photo. The motivation for PSF is that styles of sketches are consistent for facial components with similar shapes. To find corresponding sketch patches for an input photo, we only need to match the facial components with similar shapes in the reference set. Since the shapes of facial components are limited, a small reference set with a few hundreds of photo-sketch pairs is often sufficient for this purpose. However, the same approach cannot be used for training the sketch-to-photo generator $G_{s2p}$ because sketch patches with the same shape may give rise to photo patches of many different styles. Instead, we follow Cycle-GAN and use sketches generated by $G_{p2s}$ to train $G_{s2p}$ in a self-supervised manner. Although the proposed PSF helps to constrain the contents of the output sketches from $G_{p2s}$, we find \emph{steganography} still exists and is quite harmful to the training of $G_{s2p}$ because it learns to cheat. To solve this problem, we employ a simple {\em noise-injection} strategy to disrupt the invisible steganography and force $G_{s2p}$ to learn better translation from sketches to photos. Although the inputs of $G_{s2p}$ are noisy during training, we observed that $G_{s2p}$ can handle clean sketches quite well during testing due to the intrinsic image prior of CNNs \citep{UlyanovVL17DIP}. Experiments demonstrated that the {\em noise-injection} strategy can largely benefit the training of $G_{s2p}$.

In summary, our main contributions are:
\begin{itemize}
  \item We propose a semi-supervised learning framework based on Cycle-GAN, named Semi-Cycle-GAN, for face photo-sketch translation. 
  \item The proposed {\em pseudo sketch feature} (PSF) allows us to train $G_{p2s}$ using a small reference set of photo-sketch pairs together with a large face photo dataset without ground-truth sketches. This enables our networks to generalize well to face photos in the wild.
  \item We introduce a self-supervised approach to train the sketch-to-photo generator $G_{s2p}$ \emph{without using real sketches} through cycle-consistency. In particular, we find that cycle-consistency loss suffers greatly from invisible steganography, and the simple \emph{noise-injection} strategy helps a lot to improve it. 
\end{itemize}

A preliminary version of this work appeared in \cite{chen2018face-sketch-wild}. We extend it in five aspects: (1) we combine our previously proposed semi-supervised learning framework with cycle-consistency to conduct both photo-to-sketch and sketch-to-photo translations; (2) we find that cycle-consistency loss suffers greatly from invisible steganography, and the simple \emph{noise-injection} strategy helps a lot to improve it; (3) we add a Gram matrix loss based on PSF which provides second-order style supervision; (4) we provide more comparisons with recently proposed methods such as PS2MAN \citep{Wang2017psman}, SCA-GAN \citep{Gao2017cagan}, Knowledge Transfer \citep{Zhu2019KT} (denoted as KT), GENRE \citep{li2021high} and PANet \citep{nie2022panet}; (5) we adopt recent perceptual oriented metrics (\ie, LPIPS~\citep{zhang2018perceptual}, DISTS~\citep{ding2020iqa}, and FID~\citep{heusel2017fid}) for performance evaluation. In particular, our extended framework shows better performance than \cite{chen2018face-sketch-wild}.

\section{Related Works} \label{sec:related-works}

\para{Exemplar-Based Methods} Since photos and sketches are in two different modalities, it is not straightforward to learn a direct mapping between them. \cite{tang2003face} introduced eigentransformation to perform exemplar matching between photos and sketches by assuming a linear transformation between them. 
\cite{liu2005nonlinear} noticed that the linear assumption holds better locally, and proposed the patch-based local linear embedding (LLE). \cite{wang2009face} introduced a multi-scale markov random fields (MRF) model to resolve inconsistency between adjacent patches. \cite{zhang2010lighting} extended MRF with shape priors and SIFT features. \cite{zhou2012markov} proposed the markov weight fields (MWF) model to synthesize new sketch patches that are not present in the training dataset. \cite{gao2012face} proposed to adaptively determine the number of candidate patches by sparse representation. \cite{wang2013transductive} proposed a transductive model which optimizes the MRF-based photo-to-sketch and sketch-to-photo models simultaneously. A few works such as \cite{song2014real}and \cite{wangrslcr} tried to improve the efficiency of the sketch generation procedure. Recent methods \cite{ijcai2017-500} and \cite{chen2017pcf} used features from a pretrained CNN network as the patch feature to replace unrobust traditional features. 

\para{Learning-Based Methods} In recent years, CNN based methods have become the mainstream. \cite{zhang2015end} proposed to directly translate the input photo to sketch with a fully convolution network (FCN). \cite{zhang2017content} introduced a branched fully convolutional network (BFCN) which is composed of a content branch and a texture branch with different losses. \cite{wang2017data} improved the vanilla GAN with multi-scale structure for face photo-sketch translation. \cite{Wang2017psman} introduced multi-scale discriminators to Cycle-GAN. \cite{zhang2018madl} proposed multi-domain adversarial learning in the latent feature space of faces and sketches. \cite{FANG-IACycleGAN} introduced VGG-based feature identity loss to better preserve identity information. \cite{Gao2017cagan} extended Cycle-GAN~\citep{CycleGAN2017} with facial parsing map and proposed the SCA-GAN. Some recent popular works \citep{yi2019apdrawinggan,yi2020line,yi2020unpaired,huang2021multi,li2020staged} consider a different kind of portrait style with simple thick lines and achieve pleasant results. However, it is out of the scope of this paper and hence we do not compare with them in this work. 

\section{Semi-Cycle-GAN with noise-injection} \label{sec:method}

\begin{figure}[t]
    \centering
    \includegraphics[width=.9\linewidth]{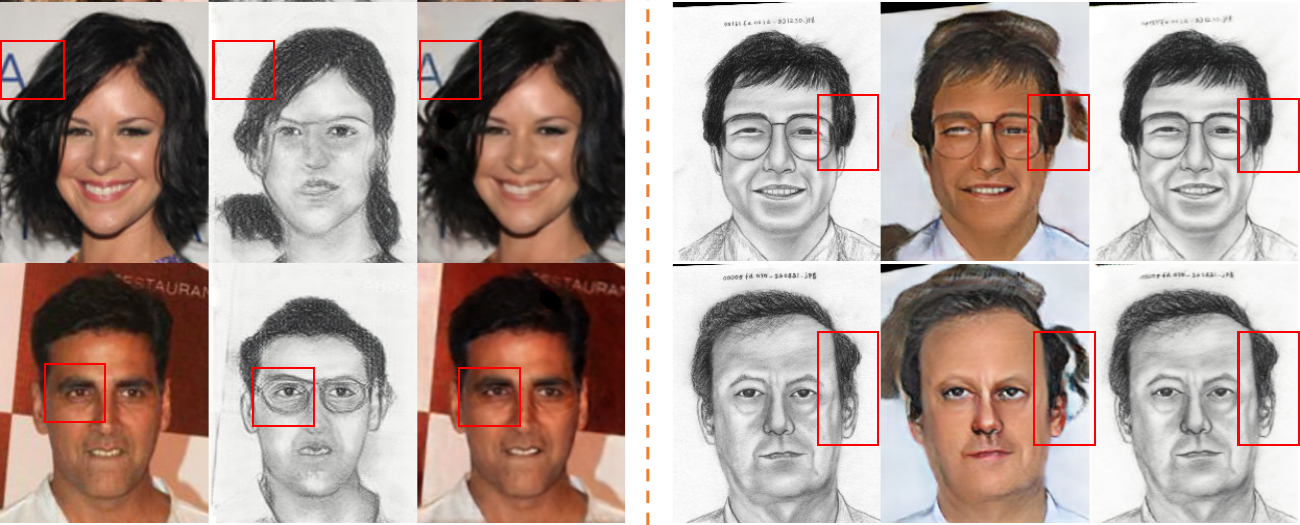}
    \\
    \makebox[0.45\linewidth]{\scriptsize Photo $\rightarrow$ Sketch $\rightarrow$ Photo }
    \makebox[0.45\linewidth]{\scriptsize Sketch $\rightarrow$ Photo $\rightarrow$ Sketch }
    \\
    \caption{Illustration of steganography when training Cycle-GAN with unpaired data.}
    \label{fig:footprint}
\end{figure}

\subsection{Steganography in Cycle-GAN} \label{sec:review-cyclegan}

\begin{figure}[!t]
  \centering
  \subfloat[Unpaired CycleGAN architecture]{\label{fig:cycle-gan}\includegraphics[width=0.45\linewidth]{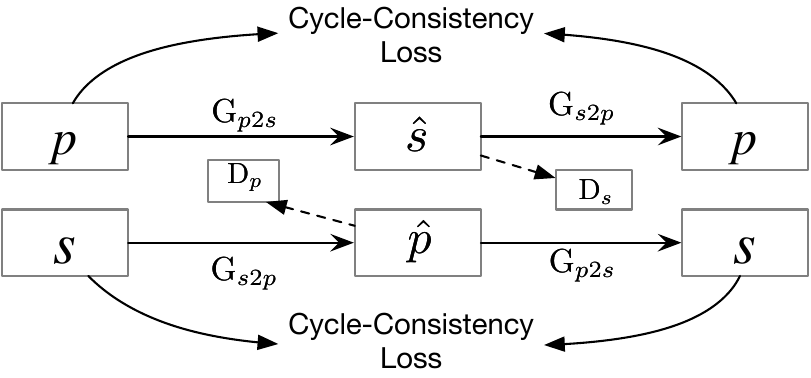}}
  \hspace{.1cm}
  \subfloat[Our Semi-Cycle-GAN architecture{\label{fig:semi-cycle-gan}}]{\includegraphics[width=0.45\linewidth]{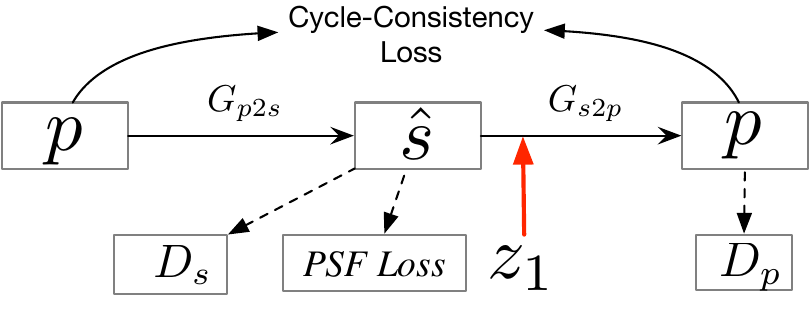}}
  \caption{Framework of unpaired Cycle-GAN and our Semi-Cycle-GAN for face-sketch translation.} \label{fig:network-arch}
\end{figure}

\begin{figure}[!t]
    \centering
    \includegraphics[width=\linewidth]{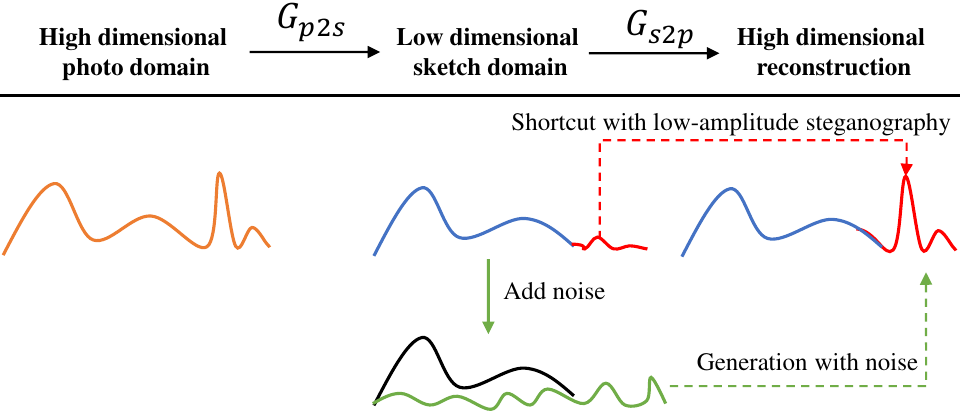}
    \caption{Theoretical illustration of how noise-injection works.}
    \label{fig:analysis_noise}
\end{figure}

In this section, we first give a brief review of the unpaired Cycle-GAN for face photo-sketch translation. We then show how Cycle-GAN cheats with invisible steganography. Given a photo set $P$ and a sketch set $S$, Cycle-GAN learns two generators: a photo-to-sketch generator $G_{p2s}$ that maps photo $p\in P$ to sketch $s\in S$, and a symmetric sketch-to-photo generator $G_{s2p}$ that maps sketch $s \in S$ to photo $p \in P$ (see \cref{fig:network-arch}(a)). Two discriminators $D_s$ and $D_p$ are used to minimize the style differences between the generated and real sketches (\ie, $\hat{s}$ and $s$) and between generated and real photos (\ie, $\hat{p}$ and $p$). Cycle-consistency losses are used to constrain content information in photo-sketch translation and are given by:
\begin{gather}
\begin{split}
L_{cyc_p} &= \mathbb{E} [\|G_{s2p}(G_{p2s}(p)) - p\|], \\
L_{cyc_s} &= \mathbb{E} [\|G_{p2s}(G_{s2p}(s)) - s\|]. 
\end{split}\label{equ:cycle-consistency}
\end{gather}
Note that \cref{equ:cycle-consistency} does not impose a direct constraint over $G_{p2s}(p)$ and $G_{s2p}(s)$, and this leads to a large solution space. \cite{chu2017steganography} pointed out that Cycle-GAN tends to hide invisible steganography in the outputs to satisfy the cycle-consistency constraint when two domains have different complexity. Specifically, in face photo-sketch translation, the photo domain $P$ is much more complex than the sketch domain $S$, which makes learning of $G_{s2p}$ much more difficult than $G_{p2s}$. As a consequence, when we train $G_{s2p}$ and $G_{p2s}$ in an unpaired manner with cycle-consistency, the networks tend to learn a trivial solution by cheating with steganography rather than learning the desired translation networks. \Cref{fig:analysis_noise} provides a theoretical illustration of steganography effect and how noise-injection helps to solve this problem. Given that the high-dimensional photo domain $P$ contains a more extensive range of information in comparison to the low-dimensional sketch domain $S$, it poses a considerable challenge for the $G_{s2p}$ network to reconstruct the missing information (\eg, hair color) from grayscale input sketches. The networks tend to learn to conceal the extra information in a low-amplitude signal (\ie, the red curve) to facilitate seamless reconstruction of the high-dimensional signal while retaining the appearance of the sketch signal. Since steganography needs to be low-amplitude signals, it is vulnerable to disruption through the application of random noise. In addition, $G_{s2p}$ with random noise will act as a normal GAN to complement missing information in the low-dimensional sketch domain.

\Cref{fig:footprint} shows some example results when training Cycle-GAN with unpaired dataset. We can observe from the left half of \cref{fig:footprint} (\emph{photo$\rightarrow$sketch$\rightarrow$photo}) that the lost letter in the generated sketch was recovered in the reconstructed photo, and extra glasses in the sketch were removed. A similar phenomenon also appears in the right half (\emph{sketch$\rightarrow$photo$\rightarrow$sketch}). Closely related works including \cite{chu2017steganography} and \cite{selfdefense} focus on how to avoid adversarial attack that is usually invisible in the images. We, on the other hand, are the first to study the visual effects brought by such steganography in face photo-sketch translation, which have been ignored by previous works based on Cycle-GAN \citep{Gao2017cagan,Wang2017psman}. 

To solve this problem, we propose the Semi-Cycle-GAN framework for face photo-sketch translation. As shown in \cref{fig:network-arch}(b), our framework is composed of four networks, namely $G_{p2s}$, $G_{s2p}$, $D_s$, and $D_p$. Unlike Cycle-GAN, we do not use the bidirectional cycle-consistency loss as a content constraint. We use PSF loss (see \cref{sec:psf} for details) to supervise the training of $G_{p2s}$, and cycle-consistency loss with {\em noise-injection} to supervise the training of $G_{s2p}$. In this manner, we can train our Semi-Cycle-GAN using a small paired photo-sketch dataset together with a large face dataset.

\subsection{Pseudo Sketch Feature} \label{sec:psf}


Given a test photo $p$, our target is to construct a pseudo sketch feature $\mr{\Phi}'(p)$ as the supervision using the reference set $\mc{R}\{(p^\mc{R}_i, s^\mc{R}_i)\}_{i=1}^N$, where $p^\mc{R}_i$ and $p^\mc{R}_i$ are photos and sketches in $\mc{R}$. We first use a pretrained VGG-19 network to extract a feature map for $p$ at the $l$-th layer, denoted as $\mr{\Phi}^l(p)$. Similarly, we can get the feature maps for photos and sketches in the reference dataset, \ie, $\{\mr{\Phi}^l(p^\mc{R}_i)\}_{i=1}^N$ and $\{\mr{\Phi}^l(s^\mc{R}_i)\}_{i=1}^N$. The feature maps are then subdivided into $k\times k$ patches for the following feature patch matching process. For simplicity, we denote a vectorized representation of a $k \times k$ patch centered at a point $j$ of $\mr{\Phi}^l(p)$ as $\mr{\Psi}_j\left( \mr{\Phi}^l(p)\right)$, and the same definition applies to $\mr{\Psi}_j\left( \mr{\Phi}^l(p^\mc{R}_i)\right)$ and $\mr{\Psi}_j(\mr{\Phi}^l\left(s^\mc{R}_i)\right)$. For each patch $\mr{\Psi}_j\left( \mr{\Phi}^l(p)\right)$, where $j = 1, 2, \ldots, m^l$ and $m^l = (H^l - k + 1)\times(W^l - k + 1)$ with $H^l$ and $W^l$ being the height and width of $\mr{\Phi}^l(p)$, we find its best match $\mr{\Psi}_{j'}\left( \mr{\Phi}^l(p^\mc{R}_{i'})\right)$ in the reference set based on cosine distance, \ie,
\begin{equation}
  (i', j') = \argmax_{\begin{subarray}{c} i^*=1\sim N \\ j^*=1\sim m^l \end{subarray}} \frac{\mr{\Psi}_j\left( \mr{\Phi}^l(p)\right)\cdot\mr{\Psi}_{j^*}\left( \mr{\Phi}^l(p_{i^*}^\mc{R})\right)}{\norm{\mr{\Psi}_j\left( \mr{\Phi}^l(p)\right)} \norm{\mr{\Psi}_{j^*}\left( \mr{\Phi}^l(p_{i^*}^\mc{R})\right)}}.
  \label{equ:cosine_dist}
\end{equation}
Since photos and their corresponding sketches in $\mc{R}$ are well aligned, the indices of the best matching result $(i', j')$ can be used directly to find the corresponding sketch feature patch, \ie, $\mr{\Psi}_{j'}\left( \mr{\Phi}^l(s_{i'}^\mc{R})\right)$ which serves as the pseudo sketch feature patch $\mr{\Psi}'_j\left( \mr{\Phi}^l(p)\right)$. Finally, we obtain the pseudo sketch feature representation (at layer $l$) for $p$ as $\{\mr{\Psi}'_j\left( \mr{\Phi}^l(p)\right)\}_{j=1}^{m^l}$. We provide an intuitive visualization of PSF in supplementary material.  

\subsection{Loss Functions}

We train generators ($G_{p2s}$, $G_{s2p}$) and discriminators ($D_s$, $D_p$) alternatively with the following loss functions 
\begin{gather}
L_G^{total} = \lambda_p L_p + \lambda_{sty} L_{sty} + \lambda_{cyc}L_{cyc} + \lambda_{adv}(L_{G_{p2s}} + L_{G_{s2p}}),\\
L_D^{total} = L_{D_{p2s}} + L_{D_{s2p}}
\end{gather}
where $\lambda_p$, $\lambda_{sty}$, $\lambda_{cyc}$, and $\lambda_{adv}$ are trade-off weights for each loss term respectively. We describe details of each term as below.

\para{Pseudo Sketch Feature Loss}
The pseudo sketch feature loss is formulated as
\begin{equation}
    L_p(p, \hat{s}) = \sum_{l=3}^5 \sum_{j=1}^{m^l} \norm{\mr{\Psi}_j\left( \mr{\Phi}^l(\hat{s})\right) - \mr{\Psi}'_j\left( \mr{\Phi}^l(p)\right)}^2,
\end{equation}
where $l=3, 4, 5$ are relu3\_1, relu4\_1, and relu5\_1 in VGG-19, and $\hat{s}$ is the predicted sketch from $G_{p2s}$. 

\para{Style Loss} Inspired by recent style transfer methods, we include Gram Matrix loss \citep{gatys2016image} as a second-order feature loss to provide better style supervision. We first average pool features in each $k\times k$ patch for both $\mr{\Psi}_j\left( \mr{\Phi}^l(\hat{s})\right)$ and $\mr{\Psi}'_j\left( \mr{\Phi}^l(p)\right)$, resulting in   features $\mr{\psi}_l$ and $\mr{\psi}'_l$ of size $m^l\times c^l$, where $c^l$ is the channel number in $l$-th layer. We then calculate the Gram Matrix loss as
\begin{equation}
  L_{sty}(p, \hat{s}) = \sum_{l=3}^5\frac{1}{(c^lm^l)^2} \| \mr{\psi}_l^T\mr{\psi}_l - \mr{\psi}_l^{'T}\mr{\psi}'_l\|^2_2,
\end{equation}

\para{Cycle-Consistency with Noise-injection} We use the cycle-consistency loss with {\em noise-injection} as supervision, which is formulated as 
\begin{equation} 
  L_{cyc}(p) = \|G_{s2p}\left(G_{p2s}(p) + \sigma z_1\right) - p\|^2_2,
\end{equation}
where $z_1$ is randomly sampled from a normal distribution with the same dimensions as $G_{p2s}(p)$, and $\sigma$ is a hyperparameter that controls the noise level. 

\para{GAN Loss} We use the hinge loss to make the training process more stable. The objective functions of hinge loss are given by
\begin{gather}
  L_G = -\mathbb{E}[D(G(x))], \\
  L_D = \mathbb{E}[\max(0, 1 - D(y))] + \mathbb{E}[\max(0, 1 + D(G(x)))],
\end{gather}
where $x, y, D$ refer to $p, s, D_s$ when $G$ is $G_{p2s}$, and $s, p, D_p$ when $G$ is $G_{s2p}$. 

\begin{figure*}[t]
\centering
\includegraphics[width=.99\linewidth]{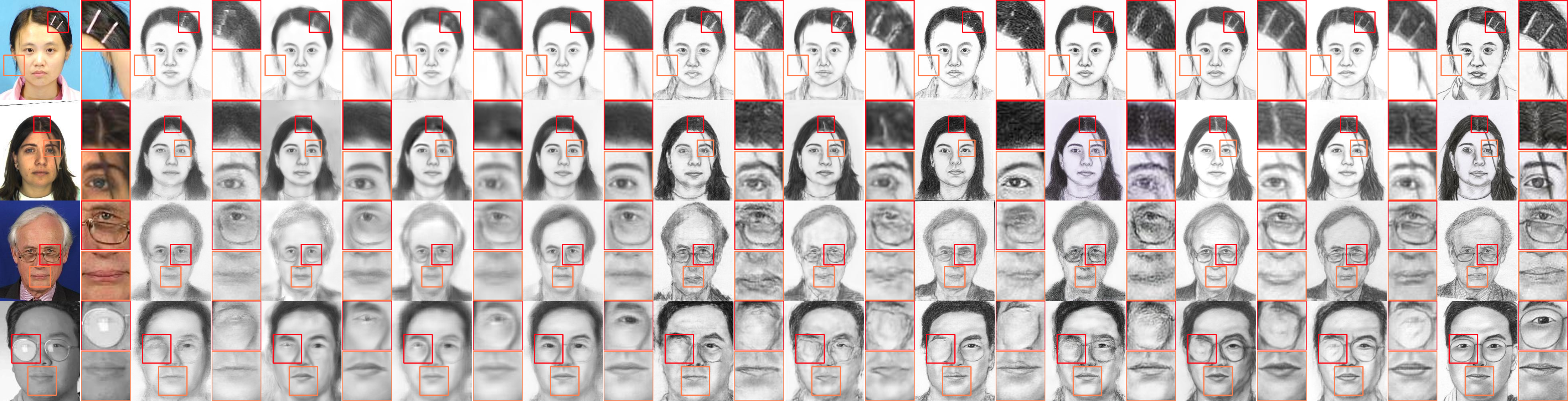} 
\scriptsize
\setlength{\tabcolsep}{0.em}
\begin{tabularx}{1.\linewidth}{*{12}{C}}
(a) Photo & (b) MWF & (c) SSD & (d) RSLCR & (e) DGFL & (f) Pix2Pix & (g) PS2MAN & (h) SCA-GAN & (i) KT & (j) FSW & (k) SCG (ours) & (l) GT Sketch
\end{tabularx}
\caption{Examples of synthesized face sketches on the CUFS dataset and the CUFSF dataset. See more examples in supplementary material.}\label{fig:result_bmk}
\end{figure*}

\section{Experiments} \label{sec:details}

\subsection{Datasets and Metrics}

\para{Datasets} To compare with previous works, we evaluate our model on two public benchmark datasets, namely the CUFS dataset (combination of CUHK~\citep{tang2003face}, AR~\citep{martinez1998r} and XM2VTS~\citep{Messer99xm2vtsdb}), and the CUFSF dataset~\citep{zhang2011coupled}. For semi-supervised learning, we use extra face photos from VGG-Face dataset~\citep{Parkhi15}. We randomly select 1,244 photos from VGG-Face to test model performance on natural images. More details are provided in supplementary material. 

\para{Training Details} We set all the trade-off weights $\lambda_p$, $\lambda_{sty}$, $\lambda_{cyc}$, and $\lambda_{adv}$ to $1$ for simplicity. We use Adam \citep{kingma2014adam} with learning rates $0.001$ for generators and $0.004$ for discriminators, and set $\beta_1=0.9, \beta_2=0.999$. The learning rates are linearly decayed to $0$ after the first 10 epochs. The training batch size is 2, and models are trained on Nvidia 1080Ti GPUs. 

\para{Metrics} For test sets with ground truth, we use FSIM~\citep{zhang2011fsim}, LPIPS~\citep{zhang2018perceptual} and DISTS~\cite{ding2020iqa} to measure the texture quality, and NLDA score to measure the identity similarity following \cite{wangrslcr}. For the evaluation of face-sketch translation in the wild, there are no ground truth sketches to calculate FSIM, LPIPS, and DISTS. We therefore exploit FID~\citep{heusel2017fid} to measure the feature statistic distance between the generated sketch datasets and real sketch datasets. We explain details of these metrics in supplementary material.

\subsection{Comparison on Public Benchmarks} \label{sec:exp_benchmark}

We evaluate our model on both photo-to-sketch and sketch-to-photo translations on CUFS and CUFSF, which were captured under laboratory settings. We compare our results both qualitatively and quantitatively with four exemplar-based methods, namely MWF~\citep{zhou2012markov}, SSD~\citep{song2014real}, RSLCR~\citep{wangrslcr}, and DGFL~\citep{ijcai2017-500}, and five GAN-based methods, namely Pix2Pix-GAN~\citep{pix2pix2017}, PS2MAN~\citep{Wang2017psman}, MDAL~\citep{zhang2018madl}, KT~\citep{Zhu2019KT} and SCA-GAN~\citep{Gao2017cagan}. We obtain the results of MWF, SSD, RSLCR, and DGFL from \cite{wangrslcr}, the results of SCA-GAN and KT from the respective authors, and use the public codes of Cycle-GAN and PS2MAN to produce the results. We also compare the photo-to-sketch translation results with our previous work FSW~\citep{chen2018face-sketch-wild}. All the models are trained on the CUFS and CUFSF datasets with the same train/test partition.  

\subsubsection{Photo-to-Sketch Translation}

\Cref{fig:result_bmk} shows some photo-to-sketch results on CUFS and CUFSF. Exemplar-based methods (\cref{fig:result_bmk}(b,c,d,e)) in general perform worse than learning-based methods (\cref{fig:result_bmk}(f,g,h,i,j,k)). Their results are over-smoothed and do not show hair textures. They also fail to preserve contents well, such as hairpins in the first row and glasses in the last row. GAN-based methods can generate better textures, but they usually produce artifacts because of the unstable training. For example, Pix2Pix produces lots of artifacts in the hair and eyes (\cref{fig:result_bmk}(f)), and PS2MAN generates lots of artifacts when the facial parts of inputs are not clear or with a strong reflection of light (see the last two rows of \cref{fig:result_bmk}(g)). Although the results of SCA-GAN look great, it suffers from incorrect parsing map guidance, such as hairpins in the first row, hairlines in the second row of  \cref{fig:result_bmk}(h). Referring to \cref{fig:result_bmk}(j,k), we have improved our previous results of FSW by introducing $L_{sty}$ and the photo reconstruction branch.   

The quantitative results with different metrics in \cref{tab:quan_p2s} support our observations. It can be observed that exemplar-based methods perform much worse in terms of all metrics including FSIM, LPIPS, DISTS and NLDA. KT shows the best FSIM score but poor perceptual scores compared with SCG. We can see from \cref{fig:result_bmk}(i) that the textures, especially hair textures, generated by KT are much worse than SCG. SCA-GAN generates better textures but the generated images might be different from the original images (\eg, missing components) due to incorrect parsing map, which also leads to poor LPIPS and DISTS scores. In contrast, our SCG presents the second best results in terms of FSIM and the best results in terms of LPIPS and DISTS. As for sketch recognition, SCG also demonstrates best NLDA score on CUFS and competitive results on CUFSF, which clearly demonstrate its superiority.

\begin{table}[t]
\caption{Quantitative results for photo-to-sketch translation. SCA-GAN$^*$ needs a parsing map as guidance.} \label{tab:quan_p2s}
\renewcommand{\arraystretch}{1.4}
\centering
\resizebox{\linewidth}{!}{
\begin{tabular}{c|cc|cc|cc|cc}
\hline
\multicolumn{1}{c|}{\multirow{2}{*}{Method}} & \multicolumn{2}{c|}{FSIM $\uparrow$} & \multicolumn{2}{c|}{LPIPS $\downarrow$} & \multicolumn{2}{c|}{DISTS $\downarrow$} & \multicolumn{2}{c}{NLDA$\uparrow$} \\ \cline{2-9} 
& CUFS & CUFSF & CUFS & CUFSF & CUFS & CUFSF & CUFS & CUFSF \\ \hline\hline
 MWF & 0.7144  & 0.7029 & 0.3671 & 0.4090 & 0.2533 & 0.2825 & 92.3 & 73.8 \\
 SSD & 0.6957 & 0.6824 & 0.4033 & 0.4283 & 0.2536 & 0.2608 & 91.1 & 70.6 \\
 RSLCR & 0.6965 & 0.6650 & 0.4042 & 0.4521 & 0.2556 & 0.2896 & 98.0 & 75.9 \\
 DGFL  & 0.7078 & 0.6957 & 0.3655 & 0.3972 & 0.2410 & 0.2480 & \underline{98.2} & \underline{78.8} \\ \hline
 Pix2Pix-GAN & 0.7153 & 0.7060 & 0.3600 & 0.3868 & 0.2151 & \underline{0.2025}  & 93.8 & 71.7 \\
 PS2MAN & 0.7157 & 0.7219 & 0.3794 & 0.4155  & 0.2430 & 0.2471 & 97.6  &  77.0 \\
 SCA-GAN$^*$ & 0.7160 & \ul{0.7268} & 0.3608 & 0.4169 & \underline{0.2005} & 0.2168 & --- & --- \\
 MDAL & 0.7275 & 0.7076 & 0.3319 & 0.3841 & 0.2037 & 0.2096 & 96.6 & 66.7 \\
 KT & \textbf{0.7369} & \textbf{0.7311} & 0.3485 & \underline{0.3743} & 0.2116 & 0.2039 & 98.0 & \textbf{80.4} \\ 
 \hline\hline
 FSW & 0.7274 & 0.7103 & \underline{0.3262} & 0.3787 & 0.2063 & 0.2111 & 98.0 & 78.04 \\
SCG (ours) & \underline{0.7343} & \underline{0.7261} & \textbf{0.3232} & \textbf{0.3489} & \textbf{0.1967} & \textbf{0.184} & \textbf{98.6} & 78.1 \\
 \hline
\end{tabular}
}
\end{table}

\subsection{Sketch-to-Photo Translation} \label{sec:compare_s2p}

\begin{table}[t]
\caption{Quantitative results for sketch-to-photo translation. SCA-GAN$^*$ needs a parsing map as guidance.}
\label{tab:quan_s2p}
\renewcommand{\arraystretch}{1.2}
\centering
\resizebox*{\linewidth}{!}{
\begin{tabular}{c|cc|cc|cc|cc}
\hline
\multicolumn{1}{c|}{\multirow{2}{*}{Method}} & \multicolumn{2}{c|}{FSIM $\uparrow$} & \multicolumn{2}{c|}{LPIPS $\downarrow$} & \multicolumn{2}{c|}{DISTS $\downarrow$} & \multicolumn{2}{c}{NLDA$\uparrow$} \\ \cline{2-9} 
& CUFS & CUFSF & CUFS & CUFSF & CUFS & CUFSF & CUFS & CUFSF \\ \hline\hline
 Pix2Pix-GAN & 0.7598 & 0.7877 & 0.3977 & 0.4025 & 0.2421 & 0.2481 & 87.1 & \underline{51.4} \\
 PS2MAN & 0.7645 & 0.7807 & 0.3668 & 0.4267 & 0.2254 & 0.2706 & 84.7 & 42.2 \\
 SCA-GAN$^*$ & 0.7633 & \textbf{0.8304} & \underline{0.3251} & \textbf{0.3198} & \underline{0.1794} & \textbf{0.1829} & --- & --- \\
 KT & \textbf{0.7794} & \underline{0.7932} & \textbf{0.3233} & 0.3758 & 0.1821 & 0.2379 & \textbf{93.8} & \textbf{65.9} \\ 
 \hline\hline
 SCG (ours) & \underline{0.7652} & 0.7777 & 0.3374 & \underline{0.3527} & \textbf{0.1710} & \underline{0.2082} & \underline{90.0} & 49.7 \\
 \hline
\end{tabular}
}
\end{table}

\begin{figure}[t]
\centering
\includegraphics[width=.99\linewidth]{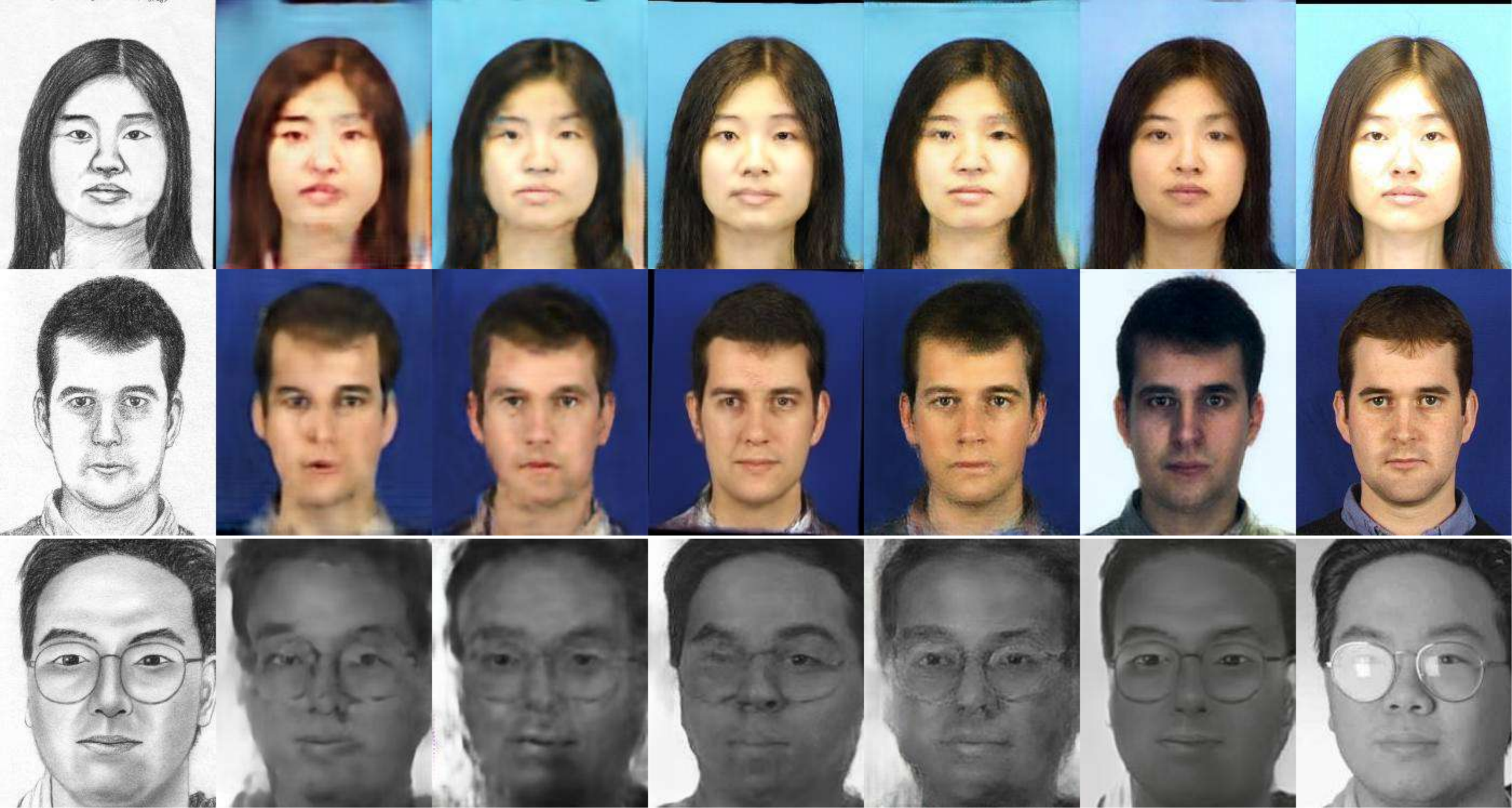} 
\setlength{\tabcolsep}{0.em}
\tiny
\begin{tabularx}{.99\linewidth}{*{7}{C}}
(a) Sketch & (b) Pix2Pix & (c) PS2MAN & (d) SCA-GAN & (e) KT & (f) SCG (ours) & (g) Photo \end{tabularx}
\caption{Examples of synthesized face photos on the CUFS dataset and the CUFSF dataset.}\label{fig:result_bmk_s2p}
\end{figure}

\Cref{fig:result_bmk_s2p} shows some example sketch-to-photo results. Same as photo-to-sketch translation, the results of Pix2Pix and PS2MAN contain many undesired artifacts. SCA-GAN produces results with the best visual quality, which is consistent with the quantitative results shown in \cref{tab:quan_s2p}. However, it still generates results with missing components under incorrect parsing map predictions, such as the missing eyes and glasses in the last row of \cref{fig:result_bmk_s2p}(d). Without any GAN losses, KT suffers from unrealistic textures. For instance, results in \cref{fig:result_bmk_s2p}(e) are grainy. Although SCG is trained in a self-supervised manner without seeing any real input sketches, it still shows competitive performance. Referring to \cref{tab:quan_s2p}, SCG shows the best or second results in 5 out of 8 columns. The biggest problem of SCG is that the synthesized colors are quite different from the ground truth. This is legitimate because the model is not suppose to recover exact color as ground truth unless overfitting.  

\subsection{Photo-to-Sketch Translation in the Wild} \label{sec:exp_wild}

\begin{figure}[t]
\centering
\includegraphics[width=0.99\linewidth]{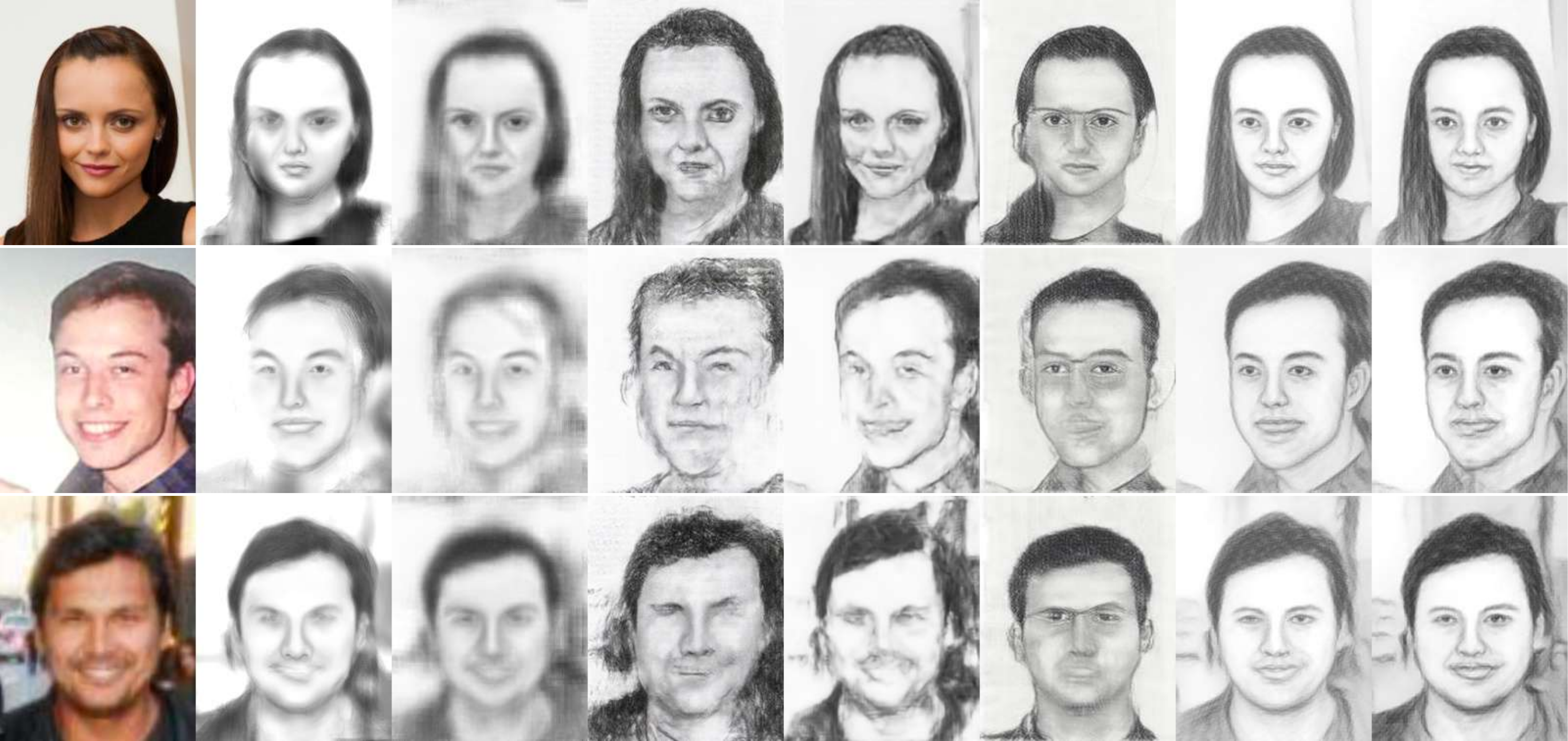}
\setlength{\tabcolsep}{0.em}
\tiny
\begin{tabularx}{.99\linewidth}{*{8}{C}}
(a) Photo & (b) SSD & (c) RSLCR & (d) Pix2Pix & (e) PS2MAN & (f) Cycle-GAN & (g) FSW & (h) SCG (ours) \end{tabularx}
\caption{Comparison for images in the wild. Benefiting from the additional training data, SCG can deal with various photos.}
   \label{fig:result-wild}
\end{figure}

\begin{table}[t]\centering
    \caption{Quantitative results and user study for photo-to-sketch translation in the wild.}\label{tab:quan_wild}
    \begin{subtable}[h]{0.39\linewidth}
      \resizebox*{0.99\linewidth}{!}{
      \begin{tabular}{c|c}
      \hline
      Method & FID$\downarrow$ \\ \hline \hline
      SSD & 94.6 \\ 
      Fast-RSLCR & 144.0  \\ 
      Pix2Pix-GAN & 86.7 \\ 
      PS2MAN & 90.8 \\ 
      Cycle-GAN & 87.8 \\ 
      FSW & \underline{81.3} \\ 
      SCG (ours) & \textbf{67.9} \\ 
      \hline
      \end{tabular}
      }
    \end{subtable}
    \begin{subtable}[h]{0.59\linewidth}
      \includegraphics[width=.99\linewidth]{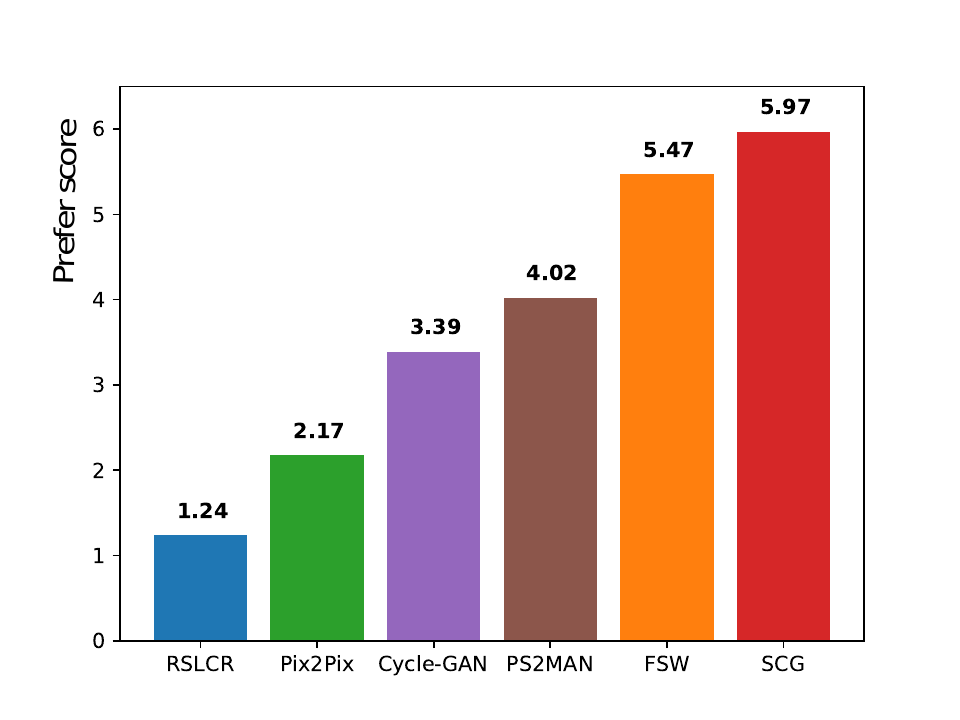}
    \end{subtable}
\end{table}

In this section, we will focus on photo-to-sketch translation in the wild. Since there are too many sketch styles in the wild, sketch-to-photo translation in the wild is beyond the scope of this paper, and we will leave it for future work. We compare SCG with other methods which provide codes, including SSD, RSLCR, Pix2Pix-GAN, PS2MAN, Cycle-GAN. \Cref{fig:result-wild} shows some photos sampled from our VGG-Face test dataset and the sketches generated by different methods. It can be observed that these photos may show very different lightings and poses \etc. Among the results of other methods, exemplar-based methods (see \cref{fig:result-wild}(b,c)) fail to deal with pose changes and different hairstyles. Although GANs can generate some sketch-like textures, none of them can well preserve the contents. The face shapes are distorted and the key facial parts are lost. It can be seen from \cref{fig:result-wild}(g,h) that only FSW and SCG can handle photos in the wild well and generate pleasant results. Compared with FSW, SCG can generate more realistic shadows and textures. The same conclusion can also be drawn from the quantitative results shown in \cref{tab:quan_wild}. We also conduct user study to better evaluate their subjective performance, as shown in \cref{tab:quan_wild} right part. We notice that our methods (FSW and SCG) are much preferred over previous methods. By introducing the $G_{s2p}$ branch and cycle-consistency, SCG further improves the performance of our previous work FSW. Details of user study are in supplementary material.  

\begin{table}[!t]
    \caption{Quantitative comparison on WildSketch dataset.}
    \label{tab:wildsketch}
    \centering
    \renewcommand{\arraystretch}{1.2}
    \small
    \begin{tabular}{c|*{6}{c}}
    \hline
         Method &  Cycle-GAN & GENRE & CA-GAN & PANet & Ours \\ \hline
         FSIM$\uparrow$ & 0.6654 & 0.6902 & \underline{0.6960} & 0.6950 & \textbf{0.7010} 
    \\ \hline
    \end{tabular}
\end{table}

\begin{figure}[!t]
    \centering
    \includegraphics[width=\linewidth]{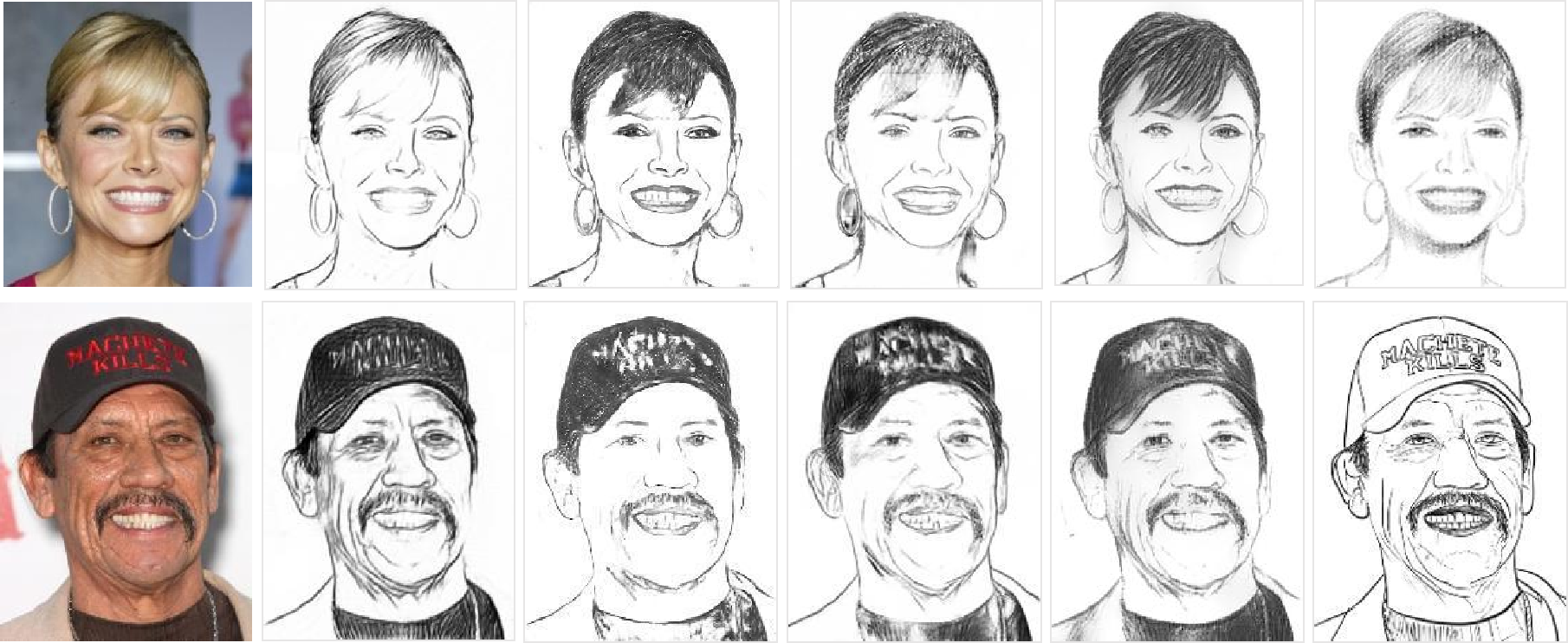}
    \setlength{\tabcolsep}{0.em}
    \tiny
    \begin{tabularx}{\linewidth}{*{6}{C}}
    (a) Photo & (b) Cycle-GAN & (c) (S)CA-GAN & (d) PANet & (e) Ours & (f) GT \end{tabularx}
    \caption{Example comparison with WildSketch dataset.}
    \label{fig:wildsketch}
\end{figure}   

We have also included comparison on the latest in-the-wild benchmark WildSketch \cite{nie2022panet}, as shown in \cref{tab:wildsketch} and \cref{fig:wildsketch}. Our findings indicate that when incorporating additional, diverse photos from the VGG face dataset, our method achieves SoTA performance. \Cref{fig:wildsketch} also supports our claim that the inclusion of extra training photos improves generalization abilities of our model. For instance, ours demonstrates greater robustness towards hair color variations in the first row and shows better result with the presence of a hat in the second row. These results underscore the effectiveness of our proposed semi-supervised approach.

\subsection{Ablation Study}

\begin{table}[t]
\caption{Ablation study of Semi-Cycle-GAN. $\sigma$: noise level, $k$: feature patch size, $L_{sty}$: use second-order style loss or not.} \label{tab:ablation_scg}
\centering
\resizebox*{\linewidth}{!}{
\begin{tabular}{*{8}{c}}
\hline
Configuration & \textsf{A} & \textsf{B} & \textsf{C} & \textsf{D} & \textsf{E} & \textsf{F} & \textsf{G} \\ \hline
$\sigma$ & 0 & 10 & \red{20} & 30 & 20 & 20 & 20 \\ \hdashline
$k$ & 1 & 1 & 1 & 1 & \red{3} & 5 & 3 \\  \hdashline
$L_{sty}$ & \xmark & \xmark & \xmark & \xmark & \xmark & \xmark & \red{\cmark} \\ \hline
LPIPS$\downarrow$(P2S) & 0.3260 & 0.3273 & 0.3277 & 0.3287 & \underline{0.3257}& 0.3273 & \textbf{0.3235} \\ \hdashline
LPIPS$\downarrow$(S2P) & 0.4273 & 0.3454 & 0.3435 & 0.3461 & \underline{0.3433}& 0.3447 & \textbf{0.3374} 
\\ \hline
\end{tabular}
}
\end{table}

\begin{figure}[t]\centering
  \begin{minipage}[b]{0.66\linewidth}
    \centering
    \includegraphics[width=.99\linewidth]{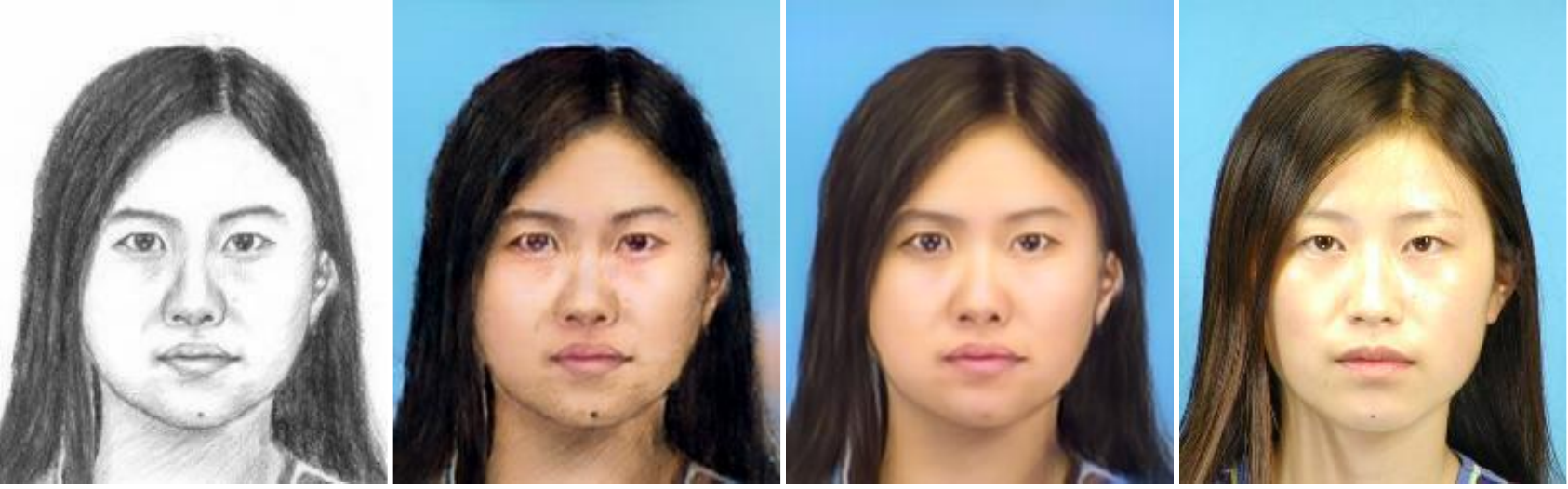}
    \tiny
    \begin{tabularx}{.99\linewidth}{*{4}{C}}
    (a) Input & (b) $\sigma=0$ & (c) $\sigma=20$ & (d) GT Photo
    \end{tabularx}
    \caption{Effect of \emph{noise-injection}.}\label{fig:ablation_scg_noise}
  \end{minipage}
  \\[0.5cm]
  \begin{minipage}[b]{0.99\linewidth}
    \centering
    \includegraphics[width=.99\linewidth]{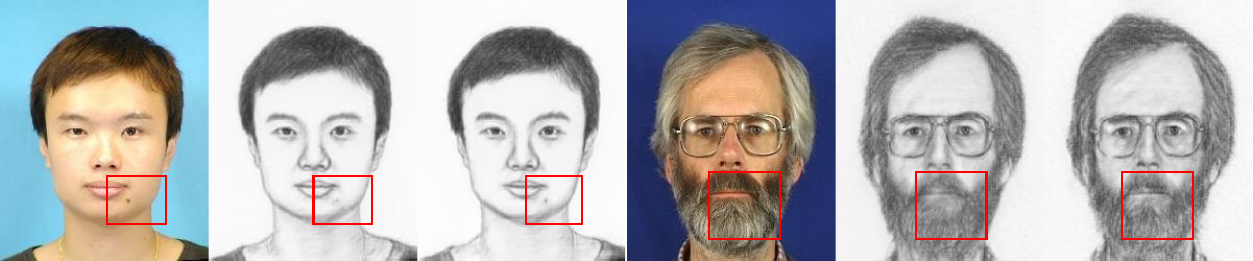}
    \scriptsize
    \begin{tabularx}{.99\linewidth}{*{6}{C}}
        Photo & w/o $L_{sty}$& w/ $L_{sty}$ 
        & Photo & w/o $L_{sty}$& w/ $L_{sty}$ 
    \end{tabularx}
    \caption{Examples of improvement on $G_{p2s}$ brought by style loss.}
    \label{fig:styleloss}
  \end{minipage}
  \\[0.5cm]
  \begin{minipage}{\linewidth}\centering
    \includegraphics[width=.99\linewidth]{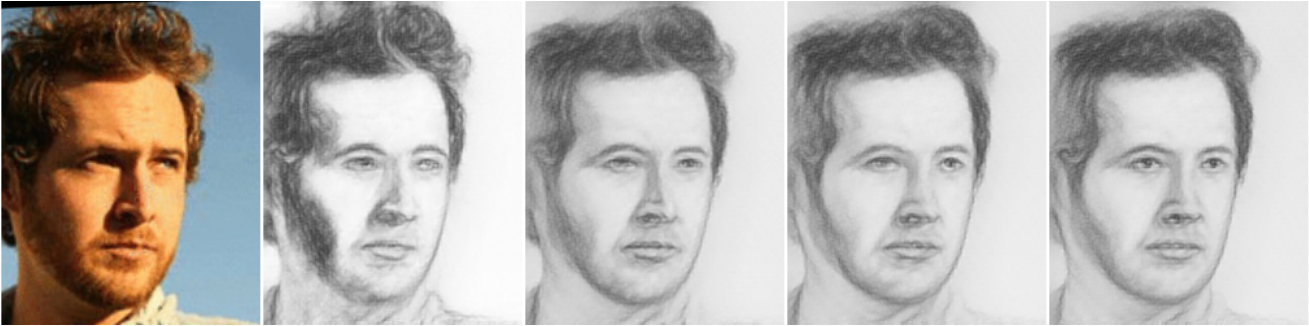}
    \\
    \makebox[0.18\linewidth]{\scriptsize Photo}
    \makebox[0.72\linewidth]{\scriptsize $\xrightarrow{\hspace*{3.5cm}\text{More extra photos}}$}
    \\
    \caption{Effectiveness of additional training photos.} \label{fig:addition-data}
  \end{minipage}
\end{figure}

To study the effectiveness of different components of the proposed method, we gradually modify the baseline Semi-Cycle-GAN and compare their results. \Cref{tab:ablation_scg} shows the results of all model variations. We discuss the results below. 

\para{Noise injection.} We show an example result with and without \emph{noise-injection} in \cref{fig:ablation_scg_noise}. It can be observed that \cref{fig:ablation_scg_noise}(c) with $\sigma=20$ is much better than \cref{fig:ablation_scg_noise}(b) with $\sigma=0$. This demonstrates that \emph{noise-injection} can greatly improve the performance of $G_{s2p}$. This is because the proposed \emph{noise-injection} strategy breaks the steganography in the outputs of $G_{p2s}$, and increases the generalization ability of $G_{s2p}$. We explore models with different levels of \emph{noise-injection}, and the results are shown in columns \textsf{A}, \textsf{B}, \textsf{C}, and \textsf{D} of \cref{tab:ablation_scg}. We can see that adding more noise is not helpful to the performance of $G_{s2p}$ but degrades the performance of $G_{p2s}$. This is likely because the backward gradients from $G_{s2p}$ are corrupted when noise-injection level is too high. We empirically find $\sigma=20$ strikes a good balance between the performance of $G_{p2s}$ and $G_{s2p}$. 

\para{Patch size.} We present the results with patch size 1, 3, and 5 in columns \textsf{C}, \textsf{E}, and \textsf{F} of \cref{tab:ablation_scg} respectively. We can observe that $k=3$ gives the best performance, while $k=5$ is worse than $k=3$. This may be caused by the fact that a large patch in the feature space represents a much larger patch in the pixel space and this leads to undesired extra contents in the pseudo sketch feature. We therefore set $k=3$ in our experiments.    

\para{Second-order style loss.} Comparing the results in columns \textsf{E} and \textsf{G} of \cref{tab:ablation_scg}, we can notice that model with $L_{sty}$ shows better performance for both $G_{p2s}$ and $G_{s2p}$. This is because $L_{sty}$ provides better style supervision for $G_{p2s}$, which can in turn benefit the training of $G_{s2p}$. \Cref{fig:styleloss} shows some examples of improvement on $G_{p2s}$ brought by style loss.

\para{Extra training photos} Introducing more training photos from VGG-Face dataset is the key to improve the generalization ability of our model. As demonstrated in \cref{fig:addition-data}, as we add more photos to the training set, the results improve significantly, see the eyes region.

\section{Conclusion} \label{sec:conclusion}
In this paper, we propose a semi-supervised Cycle-GAN, named Semi-Cycle-GAN (SCG), for face photo-sketch translation. Instead of supervising our network using ground-truth sketches, we construct a novel pseudo sketch feature representation for each input photo based on feature space patch matching with a small reference set of photo-sketch pairs. This allows us to train our model using a large face photo dataset (without ground-truth sketches) with the help of a small reference set of photo-sketch pairs. Since directly training $G_{s2p}$ in a self-supervised manner as Cycle-GAN suffers from steganography, we exploit a \emph{noise-injection} strategy to improve the robustness. Experiments show that our method can produce sketches comparable to (if not better than) those produced by other state-of-the-art methods on four public benchmarks, and outperforms them on photo-to-sketch translation in the wild.

\bibliographystyle{model2-names}
\bibliography{facesketch_bib}

\appendix
\clearpage

\section{More Methodology Details} \label{sec:method}

\subsection{Visualization of Pseudo Sketch Features}

\begin{figure*}[t]
  \captionsetup[subfigure]{labelformat=empty}
  \centering
    \subfloat[Photo \& Sketch]{\includegraphics[width=0.12\linewidth]{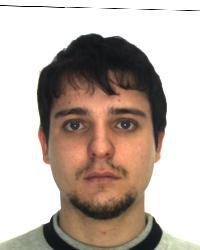}
                \includegraphics[width=0.12\linewidth]{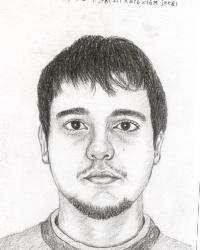}}
    \hspace{0.3cm}
    \subfloat[relu1\_1]{\includegraphics[width=0.12\linewidth]{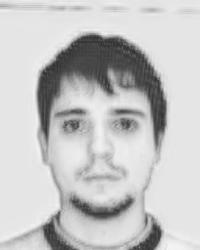}}
    \subfloat[relu2\_1]{\includegraphics[width=0.12\linewidth]{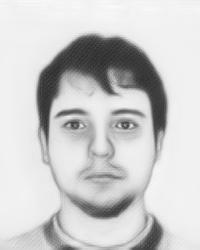}}
    \subfloat[relu3\_1]{\includegraphics[width=0.12\linewidth]{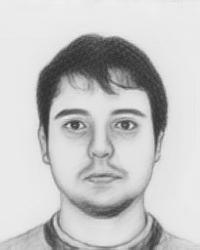}}
    \subfloat[relu4\_1]{\includegraphics[width=0.12\linewidth]{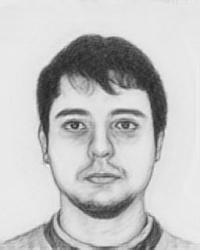}}
    \subfloat[relu5\_1]{\includegraphics[width=0.12\linewidth]{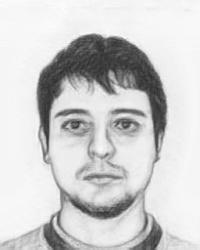}}
    \caption{Results of using different layers in pseudo sketch feature loss.}
    \label{fig:layer-example}
\end{figure*}

\begin{figure}[t]
  \centering
    \includegraphics[width=.8\linewidth]{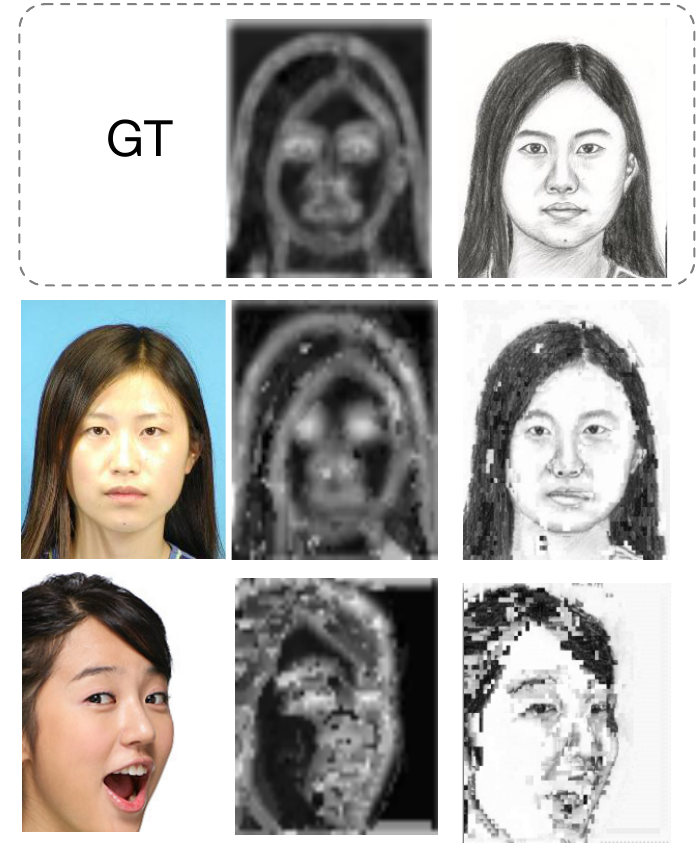}
    \begin{tabularx}{.8\linewidth}{*{3}{C}}
    Photo & PSF & Pixel Projection 
    \end{tabularx}
    \caption{Examples of PSF in the relu3\_1 layer and the pixel level projection of the patch matching result. First row: ground truth feature and sketch with the photo in second row as input. Second row: results of laboratory images. Third row: results of natural images. (\textit{Note that the pixel level results are for visualization only, and they are not actually being computed or used in training.)}}
    \label{fig:pm-example}
\end{figure}

\Cref{fig:pm-example} visualizes examples of the pseudo sketch feature. It can be seen that the pseudo sketch feature provides a good approximation of the real sketch feature (see first two columns of \cref{fig:pm-example}). We also show na\"{i}ve reconstruction obtained by directly using the matching index to index the pixel values in the reference sketches. We can see such a na\"{i}ve reconstruction does roughly resemble the real sketch, which also justifies the effectiveness of the pseudo sketch feature. Note that we only need alignment between photos and sketches in $\mc{R}$. Since we perform a dense patch matching between the input photo and the reference photos, we can also generate reasonable pseudo sketch features for input face photos under different poses (see last row of \cref{fig:pm-example}).

\subsection{Hyper-parameters of Pseudo Sketch Feature Loss}

For the pseudo sketch feature loss $L_p$, we set $l=3, 4, 5$ to relu3\_1, relu4\_1, and relu5\_1 in VGG-19. We choose these 3 layers mainly for two reasons: 1) better texture representation; 2) computation efficiency. Li \etal \cite{li2016combining} pointed out that compared with features from shallow layers, deep features after relu3\_1 are more robust to appearance changes and geometric transforms. We conduct a simple experiment to verify this, and the results are presented in \cref{fig:layer-example}. It can be observed that the model cannot synthesize sketch textures using shallow features from relu1\_1 and relu2\_1, and can generate better textures with high-level features, such as relu5\_1. However, with only high-level features, the model also generates artifacts (\eg, eyes of the sketches in \cref{fig:layer-example}). Besides, shallow feature maps have higher spatial resolutions, and it requires lots of GPU memory to calculate $L_p$. Based on the above analysis, we set $l=3, 4, 5$ to strike a balance between performance and computation cost.

\section{More Dataset and Implementation Details} \label{sec:datasets}

\begin{table}[t]
\renewcommand{\arraystretch}{1.2}
\caption{Details of benchmark datasets. (Align: whether the sketches are well aligned with photos. Var: whether the photos have lighting variations.)} \label{tab:dataset}
\centering
\resizebox*{\linewidth}{!}{
\begin{tabularx}{.99\linewidth}{c|c|c*{4}{C}}
\hline
\multicolumn{2}{c|}{Dataset} & Total Pairs  & Train & Test & Align  & Var   \\ \hline\hline
\multirow{3}{*}{CUFS} &  CUHK    & 188    & 88    & 100  & \cmark & \xmark  \\ \cline{2-7}
&  AR      & 123    & 80    & 43   & \cmark & \xmark  \\ \cline{2-7}
&  XM2VTS  & 295    & 100   & 195  & \xmark & \xmark \\ \cline{1-7}
\multicolumn{2}{c|}{CUFSF}& 1194    & 250   & 944  & \xmark & \cmark   \\ \hline
\end{tabularx}
}
\end{table}

\begin{figure}[t]
\centering
\subfloat[CUHK Student]{\includegraphics[width=0.24\linewidth]{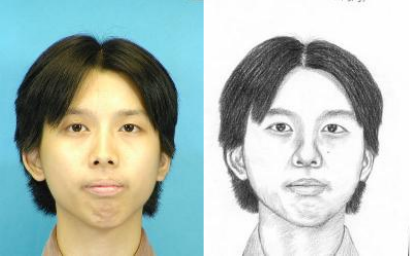}}
\subfloat[AR]{\includegraphics[width=0.24\linewidth]{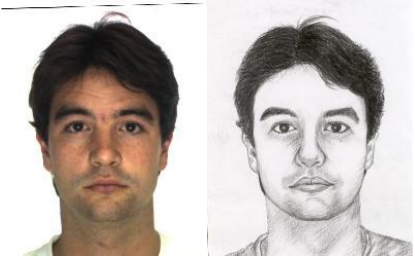}}
\subfloat[XM2VTS]{\includegraphics[width=0.24\linewidth]{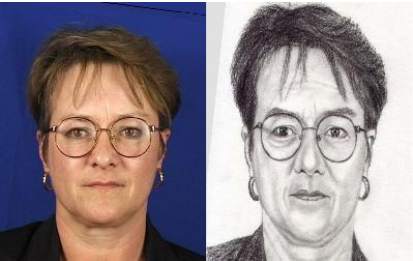}}
\subfloat[CUFSF]{\includegraphics[width=0.24\linewidth]{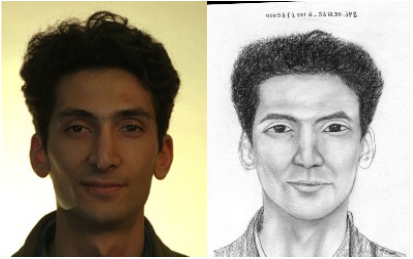}}
  \caption{Example photo-sketch pairs from existing datasets. It can be observed that (a)(b) and (c)(d) have different sketch styles (e.g., facial muscles and hair).}\label{fig:example}
\end{figure}

\paragraph{Photo-Sketch Pairs} We use four public datasets, namely the CUHK dataset~\cite{tang2003face}, the AR dataset~\cite{martinez1998r}, the XM2VTS dataset~\cite{Messer99xm2vtsdb}, and the CUFSF dataset~\cite{zhang2011coupled}, to evaluate our model. In \cite{wangrslcr,ijcai2017-500}, the first three datasets were combined to form the CUFS dataset. Note that the CUFSF dataset used in \cite{ijcai2017-500,Wang2017psman,Gao2017cagan} contains only grayscale photos. In order to train a universal model for all datasets, we collect a color version of the CUFSF dataset\footnote{Downloaded from \url{https://www.nist.gov/itl/iad/image-group/color-feret-database}} containing 986 photo-sketch pairs. Details are summarized in Table \ref{tab:dataset}, and \cref{fig:example} shows some examples of photo-sketch pairs from them. 

\begin{figure}[t]
\centering
\subfloat[Training photos]{\includegraphics[width=0.35\linewidth]{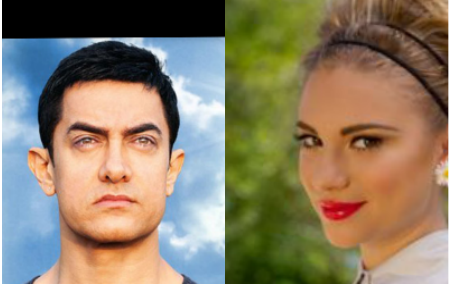}}
\subfloat[Test photos]{\includegraphics[width=0.35\linewidth]{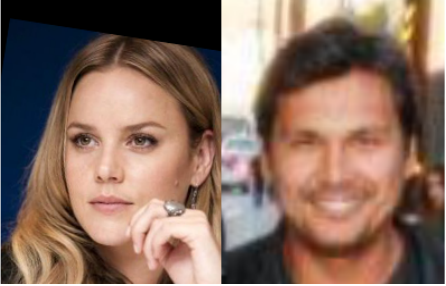}}
  \caption{Example training photos from the VGG-Face01 dataset and example test photos from the VGG test set.} \label{fig:example_vgg01}
\end{figure}

\paragraph{Face Photos} VGG-Face dataset~\cite{Parkhi15} is a popular dataset containing face photos in the wild. We use a subset of it in this work. VGG-Face has 2,622 subjects with 1,000 photos for each subject. We randomly select $\mc{N}$ photos of 2,000 subjects for training. The resulting dataset are named as VGG-Face$\mc{N}$, where $\mc{N} = 01, 02, \ldots, 10$. The VGG-Face$\mc{N}$ datasets are used to validate the performance relationship with increasing training photos. For the test dataset, 2 photos are randomly selected for each subject in the test split (no identity overlap with training dataset), which results in a VGG test set of 1,244 photos. Some examples from training and testing datasets are presented in Fig.~\ref{fig:example_vgg01}.

\paragraph{Preprocessing} For the reference datasets, we need the photo-sketch pairs to be well aligned. We perform alignment with similarity transformation based on 68 face landmarks detected using \texttt{dlib}\footnote{\url{http://dlib.net/}}. The output faces and sketches are aligned with two eyes located at $(75, 125)$ and $(125, 125)$ respectively. The output size is set to $250 \times 200 $ in order to perform a fair comparison with previous works. For photos/sketches whose landmarks cannot be detected, we simply discard them.

\subsection{Implementation Details of Patch Matching} \label{sec:implement}

Although the patch matching only happens in training stage, it is still time-consuming. We accelerate patch matching in the following three ways. First, feature patches for the photos and sketches in the reference dataset are precomputed and stored in hard disk for fast query. Second, we use a coarse-to-fine strategy to search for the best matching feature patch. We find the best-matched $n$ reference photos (we set $n=3$ in the whole training process) for the input photo based on the similarity of their relu5\_1 feature maps, which can be calculated fast. Fine-scale patch matching is then performed on these $n$ reference photos. Third, we use the convolution operator to implement Eq. (2) which can be greatly accelerated with GPU.
 
\section{More Experiment Details} \label{sec:exp}

\begin{figure*}[t]
  \centering
  \includegraphics[width=.24\linewidth]{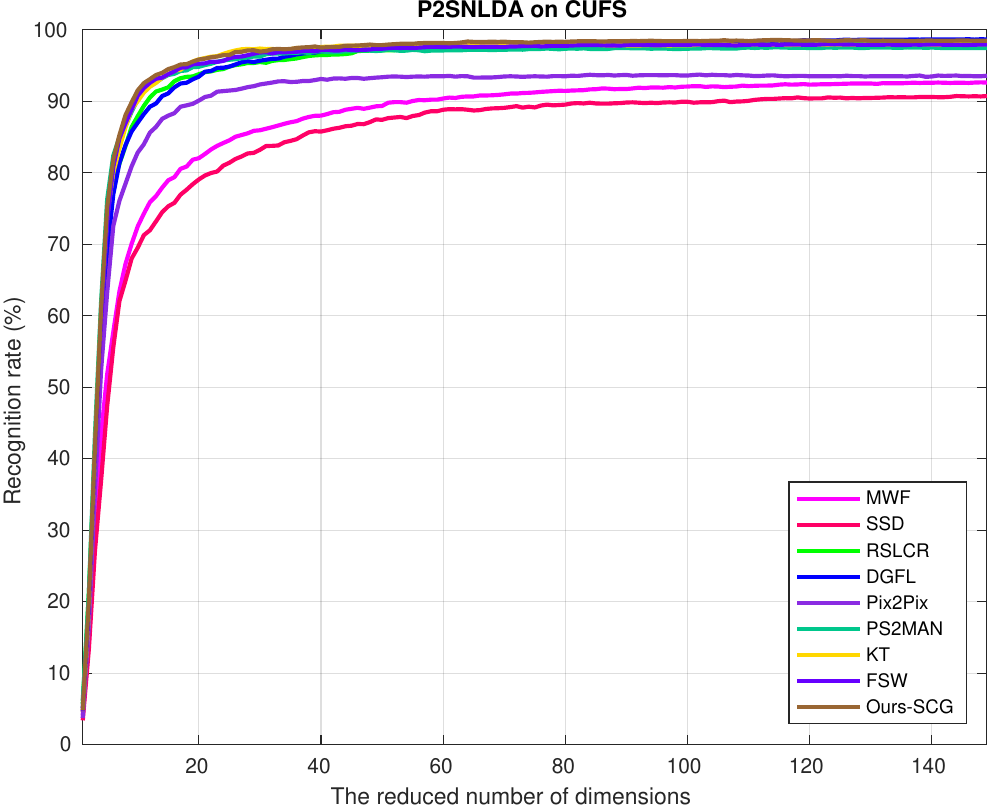}
  \includegraphics[width=.24\linewidth]{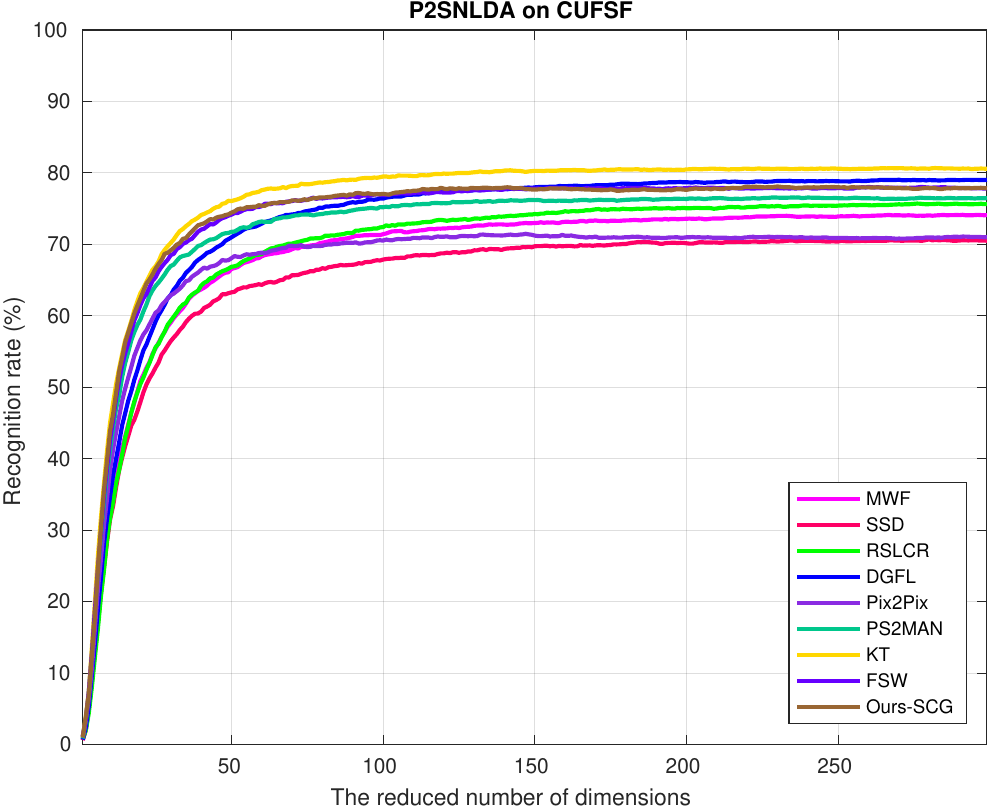}
  \includegraphics[width=.24\linewidth]{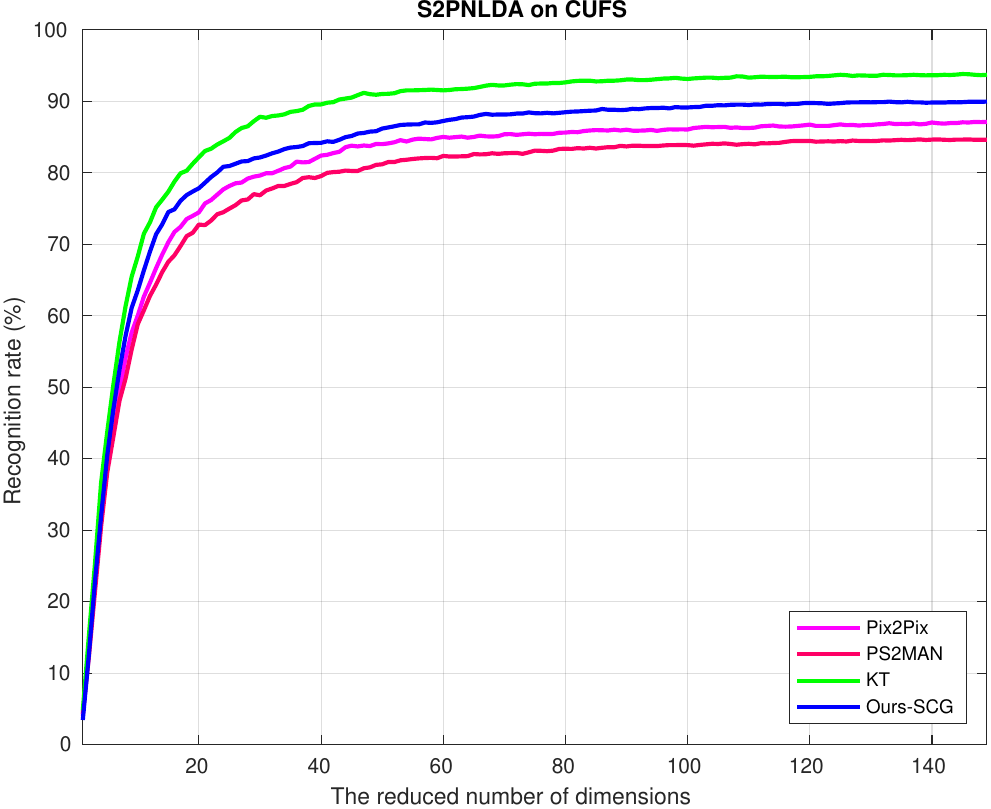}
  \includegraphics[width=.24\linewidth]{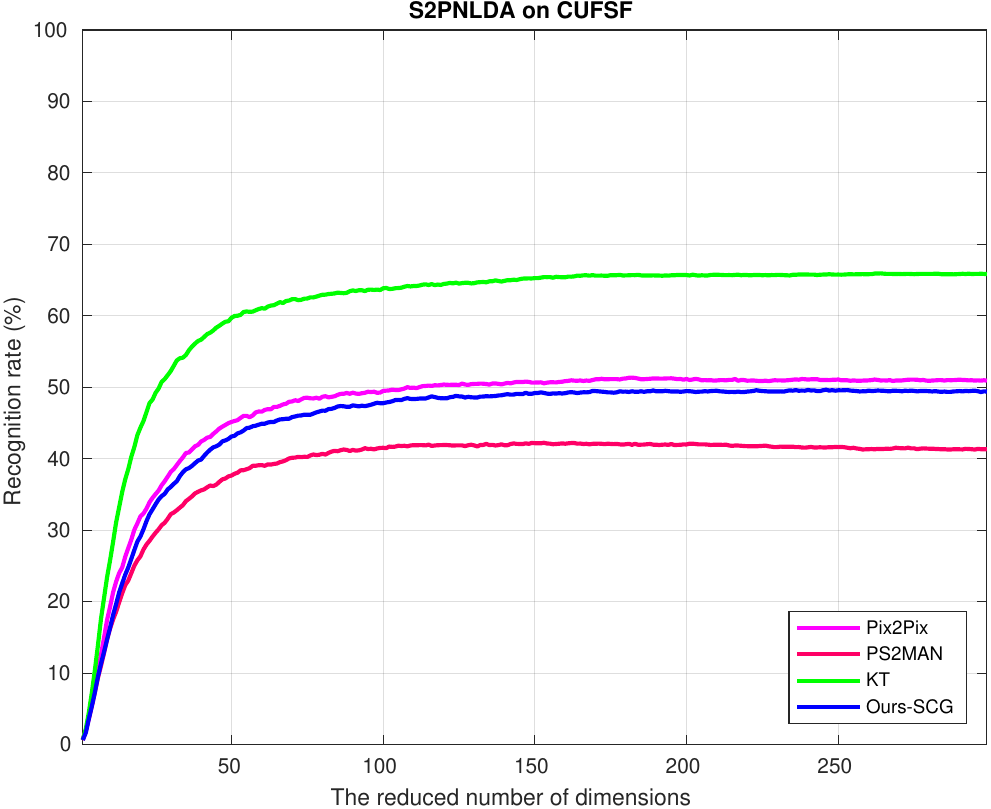}
  \scriptsize
  \begin{tabularx}{.99\linewidth}{*{4}{C}}
    (a) NLDA scores on CUFS sketches & (b) NLDA scores on CUFSF sketches & (c) NLDA scores on CUFS photos & (d) NLDA scores on CUFSF photos 
  \end{tabularx}
  \caption{Recognition rates of using synthesized face sketches (a,b) and synthesized face photos (c,d), respectively, against feature dimensions on CUFS and CUFSF.} \label{fig:nlda-score}
\end{figure*}

\subsection{Metric analysis}
\begin{figure}[t]
\includegraphics[width=1.\linewidth]{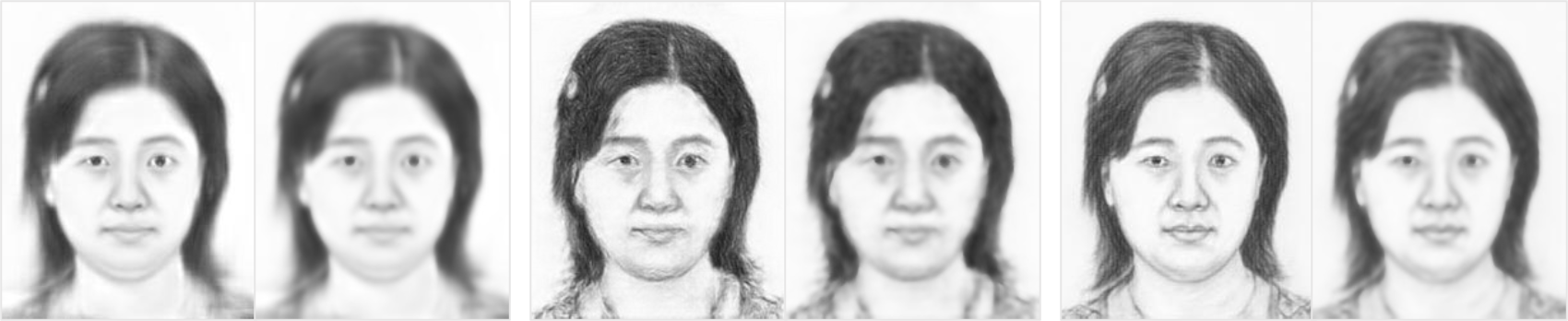}
\begin{minipage}{.32\linewidth}
\centering
\scriptsize
(a) RSLCR \\ SSIM: 0.5970/0.5903. \\   FSIM: 0.7488/0.7362.
\end{minipage}
\begin{minipage}{.32\linewidth}
\centering
\scriptsize
(b) GAN \\SSIM: 0.5648/0.5953. \\   FSIM: 0.7559/0.7506.
\end{minipage}
\begin{minipage}{.32\linewidth}
\centering
\scriptsize
(c) Ours. \\ SSIM: 0.5814/0.6055. \\   FSIM: 0.7692/0.7557.
\end{minipage}
   \caption{SSIM and FSIM scores of some generated sketches (left) and their smoothed counterparts (right).}
   \label{fig:blur-example}
\end{figure}

Previous works \cite{wangrslcr,Gao2017cagan,ijcai2017-500} usually adopted structural similarity (SSIM)~\cite{karacan2013structure} to evaluate the performance of sketch generation for test datasets with ground-truth sketches (e.g., CUFS and CUFSF). However, many works \cite{ledig2016photo,WANG2017,Wang2017psman}) pointed out that SSIM is not always consistent with the perceptual quality because SSIM favors slightly blurry images and fails to evaluate images with rich textures. To verify this, we show some sketches generated using different methods together with their SSIM scores in Fig.~\ref{fig:blur-example}. We can observe that although the results of Pix2Pix-GAN and our model have better textures, the result of RSLCR still has a better SSIM score because it is smoother. When we smooth all sketches with a bilateral filter, we notice that SSIM score for RSLCR remains almost unchanged, while the SSIM scores for Pix2Pix-GAN and our model improve by more than $1.5\%$.  

Due to the drawbacks of SSIM, we choose FSIM \cite{zhang2011fsim} as one of our image quality assessment metrics. FSIM takes local structure into account and gives lower scores to smooth results without textures, see Fig. \ref{fig:blur-example} for reference. Considering that metrics based on VGG feature space demonstrate better consistency with human perception, we also include two recent VGG-based metrics, namely LPIPS and DISTS. We use the \texttt{PyTorch} codes provided by Chen \etal \footnote{\url{https://github.com/chaofengc/IQA-PyTorch}} to calculate these metrics. For the evaluation of face-sketch translation in the wild, there are no ground truth sketches to calculate FSIM, LPIPS, and DISTS. We therefore exploit FID score to measure the feature statistic distance between the generated sketch datasets and real sketch datasets.  

\subsection{Face Recognition Details}

Following the practice of \cite{wangrslcr}, we employed  the null-space linear discriminant analysis (NLDA)~\cite{chen2000new} to perform the recognition experiments. For CUFS, we randomly selected 150 synthesized sketches and their ground-truth sketches from the test set (338 test photos) to train a classifier and used the rest 188 for testing. For CUFSF(gray, crop), we randomly selected 300 synthesized sketches and their ground-truth sketches from the test set (944 test photos) to train a classifier and used the rest 644 for testing. Each experiment was repeated 20 times. We do not include SCA-GAN in this comparison because it uses an extra face parsing map as guidance, and adopts a slightly different training/test split. 

\Cref{fig:nlda-score} shows the recognition accuracy of different methods on these two datasets. The results for photo-to-sketch translation are shown in \cref{fig:nlda-score}(a,b). We can see that the proposed SCG achieves the best NLDA scores with different feature dimensions on CUFS and competitive results on CUFSF. We notice that the recognition rate on CUFSF is similar to FSW and slightly worse than KT. This is mainly because the ground truth sketches in CUFSF are deformed too much compared with the input photos. We believe the NLDA scores on CUFSF cannot represent the quality of generated sketches. In fact, we observe that the result of SCG is clearer and without extra shadows or artifacts.

The results for sketch-to-photo translation are given in \cref{fig:nlda-score}(c,d). As mentioned in main text, it is expected that SCG cannot give the best performance because we do not use any ground-truth sketches to train $G_{s2p}$, and the colors of the results are quite different from the ground-truth photos. Nonetheless, SCG still performs better than PS2MAN on both datasets, and better than Pix2Pix on CUFS. According to the visual examples in main paper, we can observe that SCG preserves the facial components better than these two methods, and produces fewer artifacts. This explains why our results are better despite color inconsistency. 

\subsection{User Study Details}
We also conduct a user study for sketch synthesis in the wild. The methods considered include RSLCR, Pix2Pix-GAN, PS2MAN, Cycle-GAN, FSW (our previous method), and SCG (proposed). Different from our previous project which asked the subjects to rank results of different methods, we adopt a more comprehensive strategy to do the subjective study on an online human crowdsourcing platform. To be specific, we employed a two-alternative forced choice (2AFC) method. The crowd workers were shown two generated sketches at one time, and were asked to choose the sketch with better quality. An example photo-sketch pair was shown as a reference. We randomly select 30 images from VGG-Test dataset, generate sketches with the above methods and create 6 different surveys. Each survey contains results for 5 different images and $\binom{2}{6} \times 5 = 75 $ questions in total for the 6 compared methods. Each worker was asked to do one of these surveys. We collected 60 survey results from different crowd workers in total. 
We used the well-known Bradley–Terry model \cite{BradleyTerryModel} to convert the paired comparison results to global ranking. Give method $i$ and $j$, the probability that $i$ is better than $j$ is defined as
\begin{equation}
P(i > j) = \frac{e^{\beta_i}}{e^{\beta_i} + e^{\beta_j}}
\end{equation}
where $e^{\beta_{i}}$ indicates the ranking score of the method $i$. We estimate $\beta=\beta_1, \ldots,\beta_6$ by minimizing the following negative log-likelihood using gradient descent
\begin{equation}
L(\mathbf{\beta}) = - \sum_{i=1}^n \sum_{j=1, j\neq i}^n \log w_{i,j} P(i > j)
\end{equation}  
where $w_{i,j}$ indicates the total numbers that $i$ is better than $j$. The ``prefer score'' in Tab. 3 of main text refers to $e^{\beta_{i}}$.

\section{More Results} \label{sec:results}

\begin{sidewaysfigure*}[t]
  \centering
    \includegraphics[width=1.\textwidth]{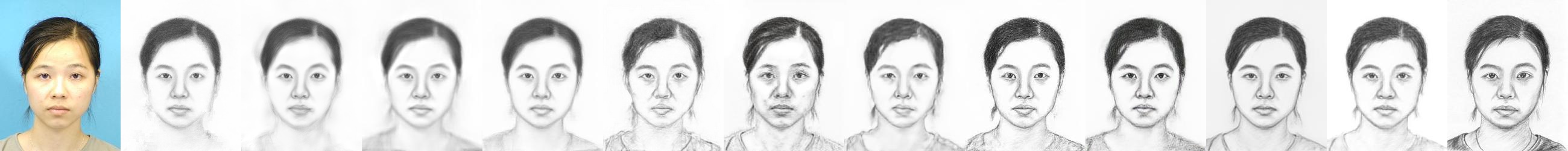}
    \includegraphics[width=1.\textwidth]{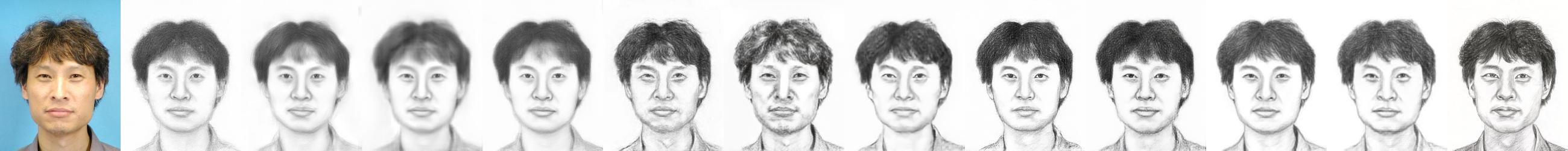}
    \includegraphics[width=1.\textwidth]{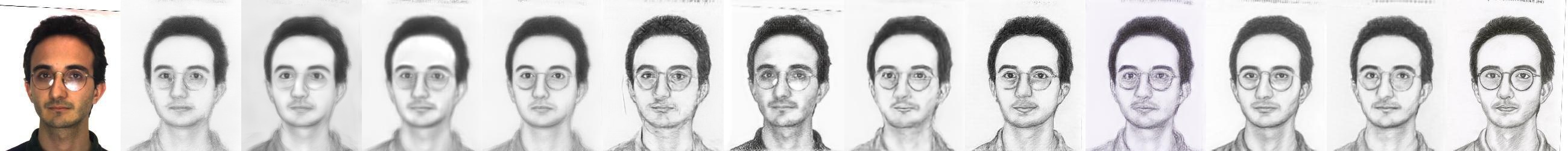}
    \includegraphics[width=1.\textwidth]{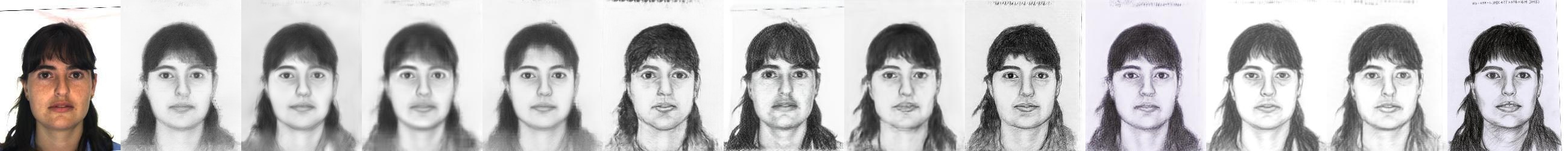}
    \includegraphics[width=1.\textwidth]{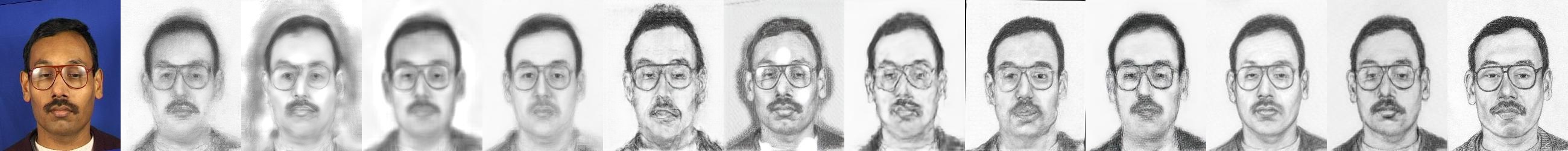}
    \includegraphics[width=1.\textwidth]{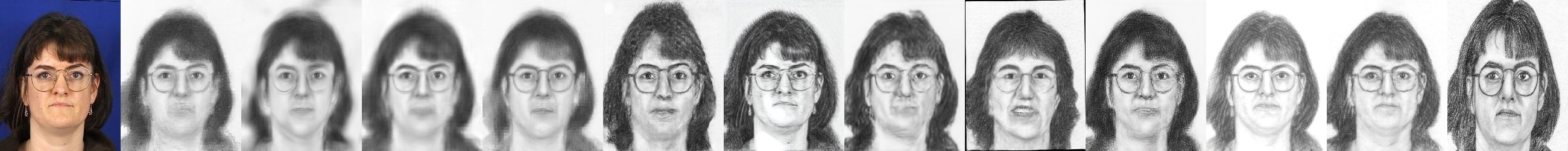}
    \begin{tabularx}{1.\textwidth}{*{13}{C}}
      (a) Photo & (b) MWF & (c) SSD & (d) RSLCR & (e) DGFL & (f) Pix2Pix & (g) Cycle-GAN & (h) PS2MAN & (i) SCA-GAN & (j) KT & (k) FSW (ours in ACCV2018) & (l) SCG (ours) & (m) GT Sketch
    \end{tabularx}
    \caption{More qualitative results for photo-to-sketch translation on CUFS test dataset.}
    \label{fig:more_results_CUFS}
\end{sidewaysfigure*}

\begin{sidewaysfigure*}[t]
  \centering
    \includegraphics[width=1.\textwidth]{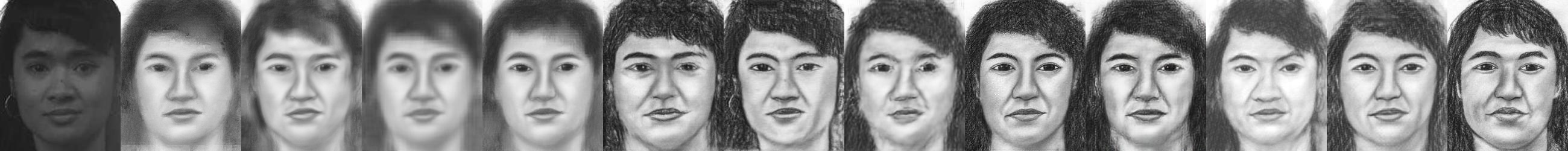}
    \includegraphics[width=1.\textwidth]{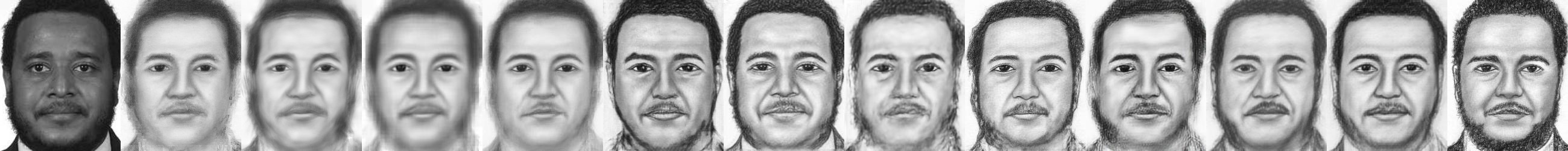}
    \includegraphics[width=1.\textwidth]{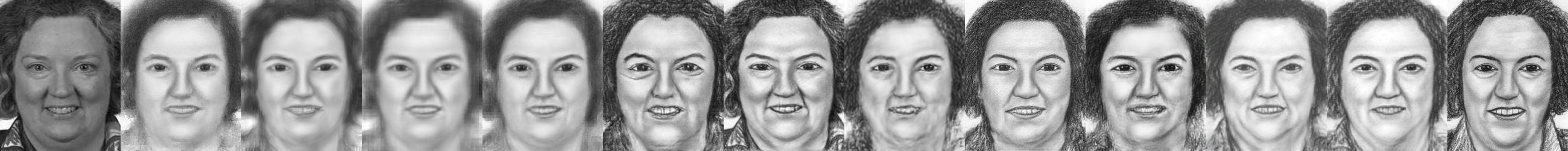}
    \includegraphics[width=1.\textwidth]{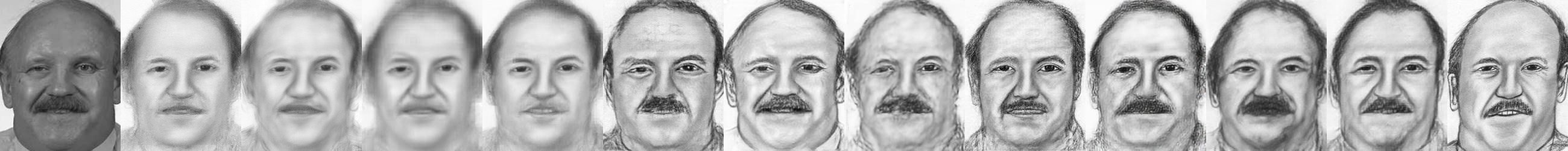}
    \includegraphics[width=1.\textwidth]{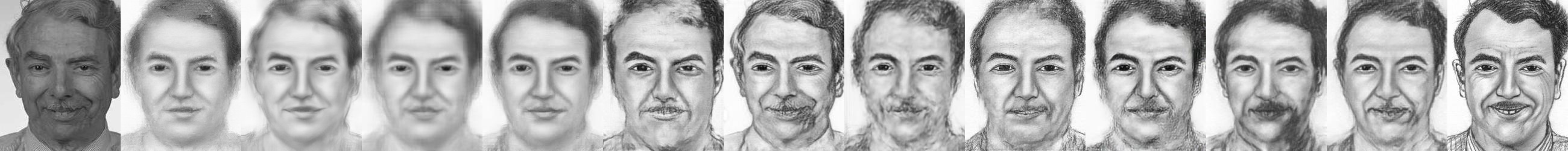}
    \includegraphics[width=1.\textwidth]{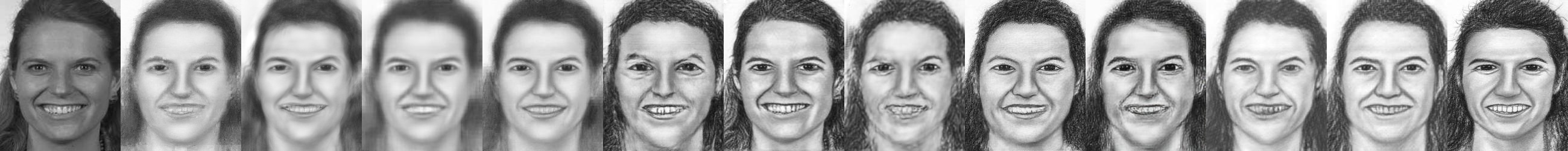}
    \begin{tabularx}{1.\textwidth}{*{13}{C}}
      (a) Photo & (b) MWF & (c) SSD & (d) RSLCR & (e) DGFL & (f) Pix2Pix & (g) Cycle-GAN & (h) PS2MAN & (i) SCA-GAN & (j) KT & (k) FSW (ours in ACCV2018) & (l) SCG (ours) & (m) GT Sketch
    \end{tabularx}
    \caption{More qualitative results for photo-to-sketch translation on CUFSF test dataset.}
    \label{fig:more_results_CUFSF}
\end{sidewaysfigure*}

In this part, we provide more photo-to-sketch results for CUFS in \cref{fig:more_results_CUFS}), CUFSF in \cref{fig:more_results_CUFSF}) and natural images of VGGFace in \cref{fig:more_result_wild}). 

\section{Limitations} \label{sec:limitations}

\begin{figure*}[htb]
\centering
\includegraphics[width=0.99\linewidth]{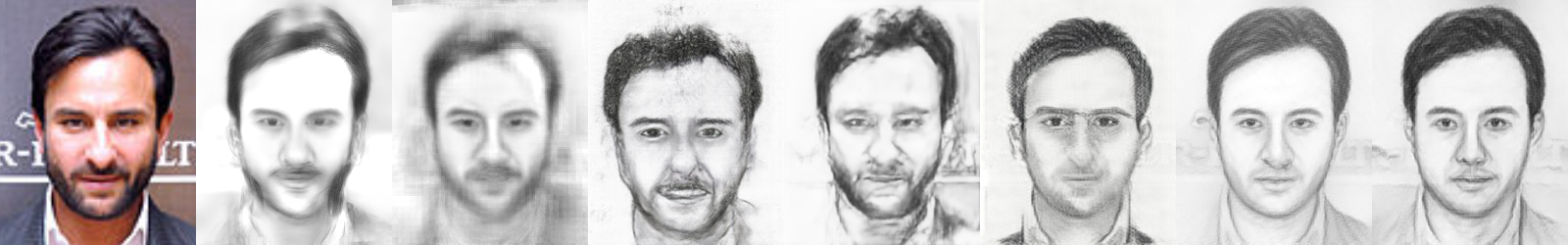}
\includegraphics[width=0.99\linewidth]{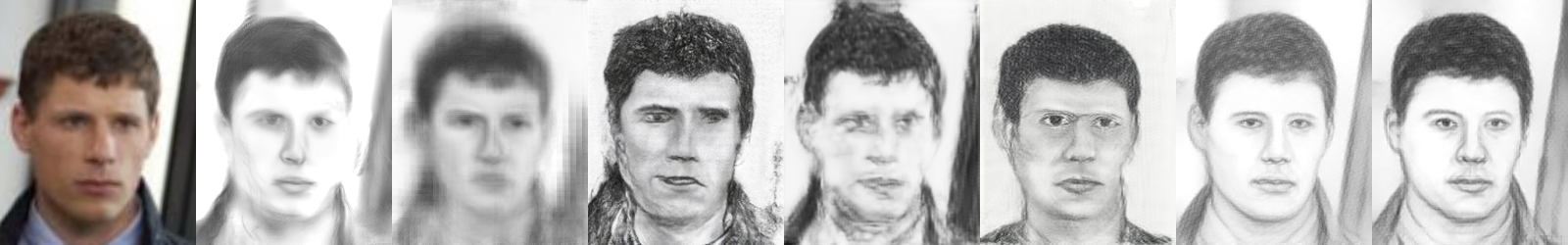}
\includegraphics[width=0.99\linewidth]{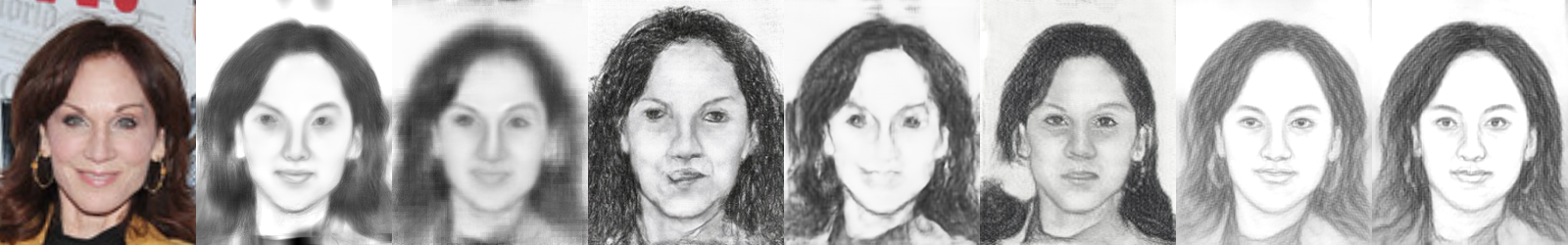}
\includegraphics[width=0.99\linewidth]{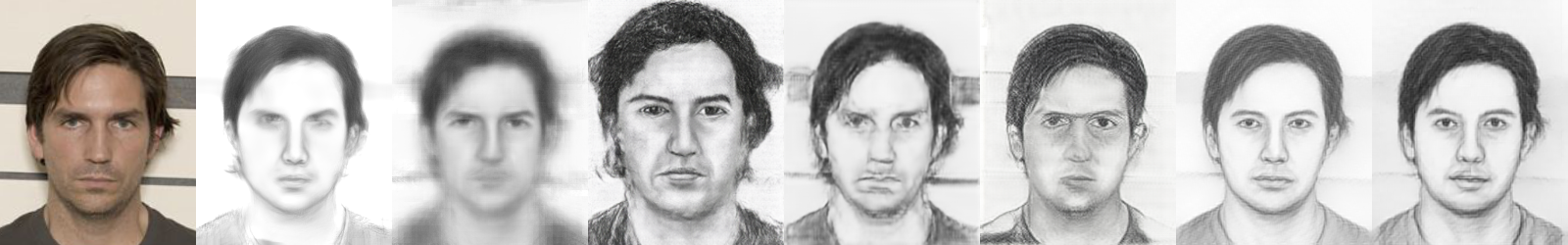}
\includegraphics[width=0.99\linewidth]{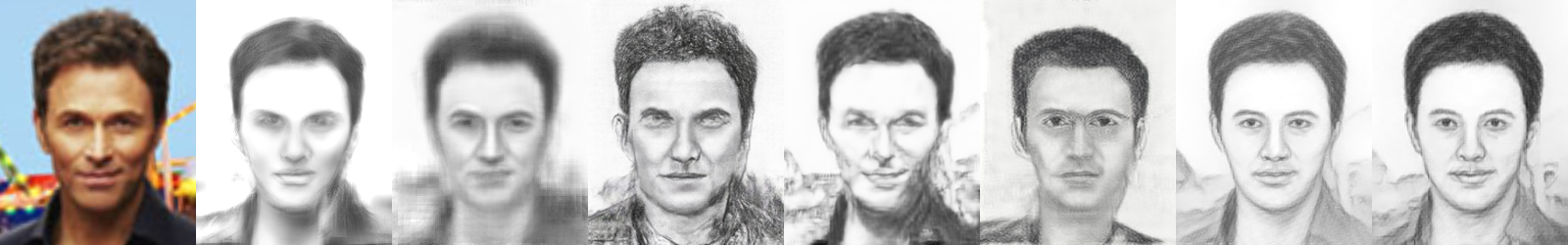}
\includegraphics[width=0.99\linewidth]{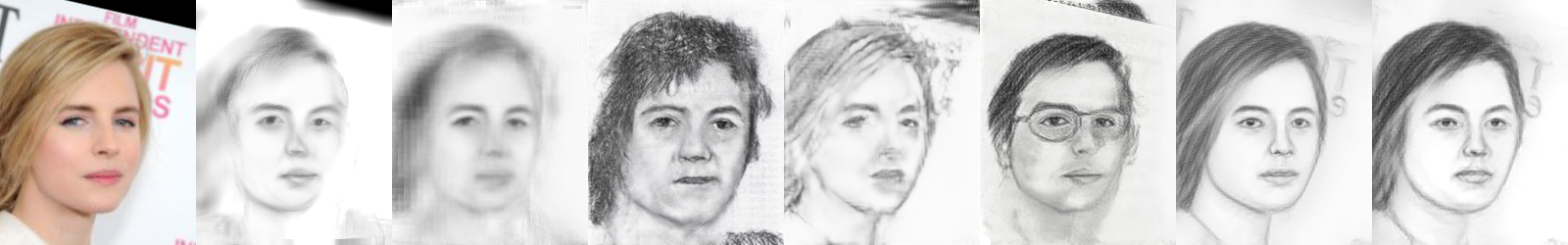}
\includegraphics[width=0.99\linewidth]{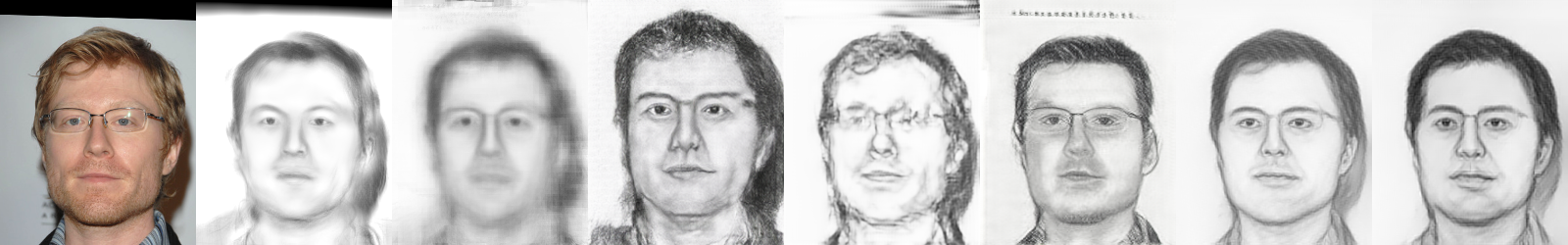}
\includegraphics[width=0.99\linewidth]{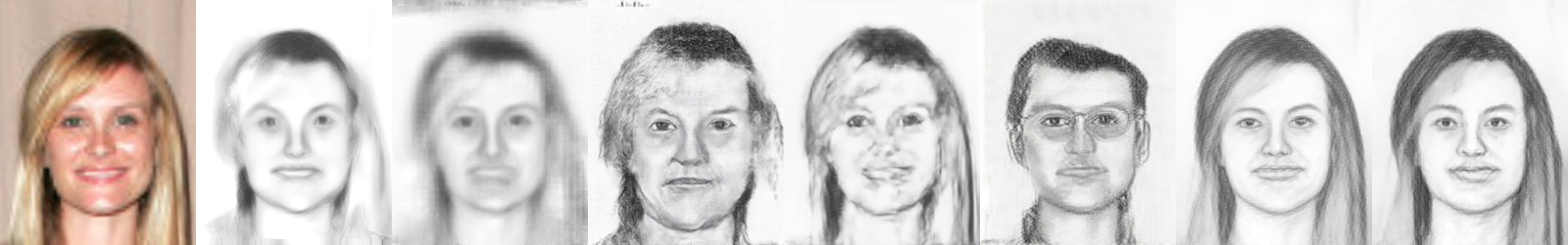}
\setlength{\tabcolsep}{0.em}
\begin{tabularx}{.99\linewidth}{*{8}{C}}
(a) Photo & (b) SSD & (c) Fast-RSLCR & (d) Pix2Pix & (e) PS2MAN & (f) Cycle-GAN & (g) FSW (ours in ACCV2018) & (h) SCG (ours) \end{tabularx}
\caption{More qualitative results for photo-to-sketch translation in the wild on VGG test dataset.}
   \label{fig:more_result_wild}
\end{figure*}

Although the proposed model shows good generalization ability to images in the wild, it cannot generate unknown structures which are not included in the small reference dataset. For example, SCG fails to generate the teeth in the last row of \cref{fig:more_result_wild}. Note that existing methods cannot even produce pleasant results for these natural images. The proposed SCG also cannot generalize to sketches with different styles, such as sketches drawn in thick lines \cite{yi2019apdrawinggan,yi2020line,yi2020unpaired}. It is quite challenging to synthesize satisfactory sketches with different styles using the same model. In a word, the above two problems are difficult to be solved with current small face sketch datasets, and we will leave them to future work.

\section{Links to public codes}

We  also provide links to the public codes used in our experiments below:
\begin{itemize}
  \item SSD: \url{http://www.cs.cityu.edu.hk/~yibisong/eccv14/index.html}
  \item RSLCR: \url{http://www.ihitworld.com/RSLCR.html}
  \item Pix2Pix-GAN: \url{https://github.com/phillipi/pix2pix}
  \item Cycle-GAN: \url{https://github.com/junyanz/pytorch-CycleGAN-and-pix2pix}
  \item PS2-MAN: \url{https://github.com/lidan1/PhotoSketchMAN}  
\end{itemize}

\clearpage

\end{document}